\documentclass[10pt,journal,compsoc]{IEEEtran}

\ifCLASSOPTIONcompsoc
  \usepackage[nocompress]{cite}
\else
  \usepackage{cite}
\fi
\usepackage{graphicx}
\usepackage{xcolor}
\usepackage{amsmath}
\usepackage{amssymb}
\usepackage{threeparttable}
\usepackage{booktabs}
\usepackage{csquotes}
\usepackage{float}
\usepackage{tabularx}
\usepackage{multirow}
\usepackage{footmisc}
\usepackage{xspace}
\usepackage{lettrine}

\newcommand{\pending}[1]{\textcolor{black}{#1}}

\newcommand{\etal}{\textit{et al}. }
\newcommand{\ie}{\textit{i}.\textit{e}. }
\newcommand{\eg}{\textit{e}.\textit{g}. }

\begin{document}
\title{A Comprehensive Survey on Deep Gait Recognition: Algorithms, Datasets, and Challenges}

\author{Chuanfu Shen,
        Shiqi Yu,~\IEEEmembership{Member,~IEEE,}
        Jilong Wang,
        George Q. Huang,
        and Liang Wang,~\IEEEmembership{Fellow,~IEEE}
\IEEEcompsocitemizethanks{
\IEEEcompsocthanksitem Chuanfu Shen is jointly with The University of Hong Kong, Hong Kong SAR, China, and Southern University of Science and Technology, China.\protect\\
E-mail: noahshen@connect.hku.hk.
\IEEEcompsocthanksitem Shiqi Yu is with the Southern University of Science and Technology, China. E-mail: yusq@sustech.edu.cn. \protect
\IEEEcompsocthanksitem Jilong Wang is with the University of Science and Technology of China and the Institute of Automation, Chinese Academy of Sciences, China. E-mail: wjltroy@gmail.com.\protect
\IEEEcompsocthanksitem George Q. Huang is with The Hong Kong Polytechnic University, Hong Kong SAR, China. E-mail: gq.huang@polyu.edu.hk.\protect
\IEEEcompsocthanksitem Liang Wang is with the Institute of Automation, Chinese Academy of Sciences, China. E-mail: wangliang@nlpr.ia.ac.cn.\protect
\IEEEcompsocthanksitem Corresponding author: Shiqi Yu.
}
}

\IEEEtitleabstractindextext{%
\begin{abstract}
Gait recognition aims to identify a person at a distance, serving as a promising solution for long-distance and less-cooperation pedestrian recognition. Recently, significant advancements in gait recognition have achieved inspiring success in many challenging scenarios by utilizing deep learning techniques. Against the backdrop that deep gait recognition has achieved almost perfect performance in laboratory datasets, much recent research has introduced new challenges for gait recognition, including robust deep representation modeling, in-the-wild gait recognition, and even recognition from new visual sensors such as infrared and depth cameras. Meanwhile, the increasing performance of gait recognition might also reveal concerns about biometrics security and privacy prevention for society. We provide a comprehensive survey on recent literature using deep learning and a discussion on the privacy and security of gait biometrics. This survey reviews the existing deep gait recognition methods through a novel view based on our proposed taxonomy. The proposed taxonomy differs from the conventional taxonomy of categorizing available gait recognition methods into the model- or appearance-based methods, while our taxonomic hierarchy considers deep gait recognition from two perspectives: deep representation learning and deep network architectures, illustrating the current approaches from both micro and macro levels. We also include up-to-date reviews of datasets and performance evaluations on diverse scenarios. Finally, we introduce privacy and security concerns on gait biometrics and discuss outstanding challenges and potential directions for future research.
\end{abstract}

\vspace{-0.5em}
}

\maketitle

\IEEEdisplaynontitleabstractindextext
\IEEEpeerreviewmaketitle

\IEEEraisesectionheading{\section{Introduction}}

\IEEEPARstart{G}{ait} recognition is a biometrics application that aims to identify pedestrians by their walking patterns~\cite{usfgait,nixon1996automatic}. It can be viewed as a vision-based person retrieval problem with the objective of human identification from gait sequences captured by visual cameras. One significant advantage of gait recognition is its ability to perform human identification at a distance, making it suitable for low-resolution and long-distance scenarios~\cite{nixonbook}. In the context of the COVID-19 pandemic~\cite{covid}, where traditional surveillance systems relying on face, iris, or fingerprint may be deficient. Therefore gait recognition emerges as a preferred solution.
Additionally, gait recognition offers the benefit of requiring less active cooperation from individuals.
In conclusion, these unique characteristics make gait recognition significantly potential for diverse applications, including surveillance, forensics, and healthcare.

Although research on gait recognition shortly started three decades ago~\cite{niyogi1994analyzing,nixon1996automatic}, the field of study has continuously advanced and expanded over the years. To our best knowledge, we categorize the evolution of gait recognition research into three stages. \textit{The initial stage}, in the early 1990s~\cite{niyogi1994analyzing}, focused on exploring the feasibility of human identification at a distance. Although the early approaches showed reasonable performance, they were evaluated on small-scale benchmarks with a limited number of subjects, typically ten at most~\cite{soton2002,usfgait}.

\textit{The second stage} emerged from DARPA Human Identification at a Distance (HumanID) program~\cite{HumanID,humanID1,humanID2}, which not only promoted techniques but also introduced valuable datasets. At this time, the methods gradually formed two categories: appearance-based and model-based. The appearance-based methods~\cite{han2005gei,CGItemplate,MEI_Template} directly exploit shape information from gait representations like silhouettes, unlike model-based methods~\cite{TraditionalModel1,TraditionalModel2,TraditionalModel3} that explicitly model a deformable human body to represent individuals. Additionally, datasets began to include over a hundred subjects~\cite{casiab,dataset_soton_large,dataset_casiac}, and they started to explore factors like view variants~\cite{usfgait,umd} and appearance-changing scenarios~\cite{casiab,nixonbook}. With the promising evaluation performance in this stage, gait recognition demonstrated the feasibility and potential for further exploration.

\begin{figure*}[ht]
\centering
\includegraphics[width=0.92\textwidth]{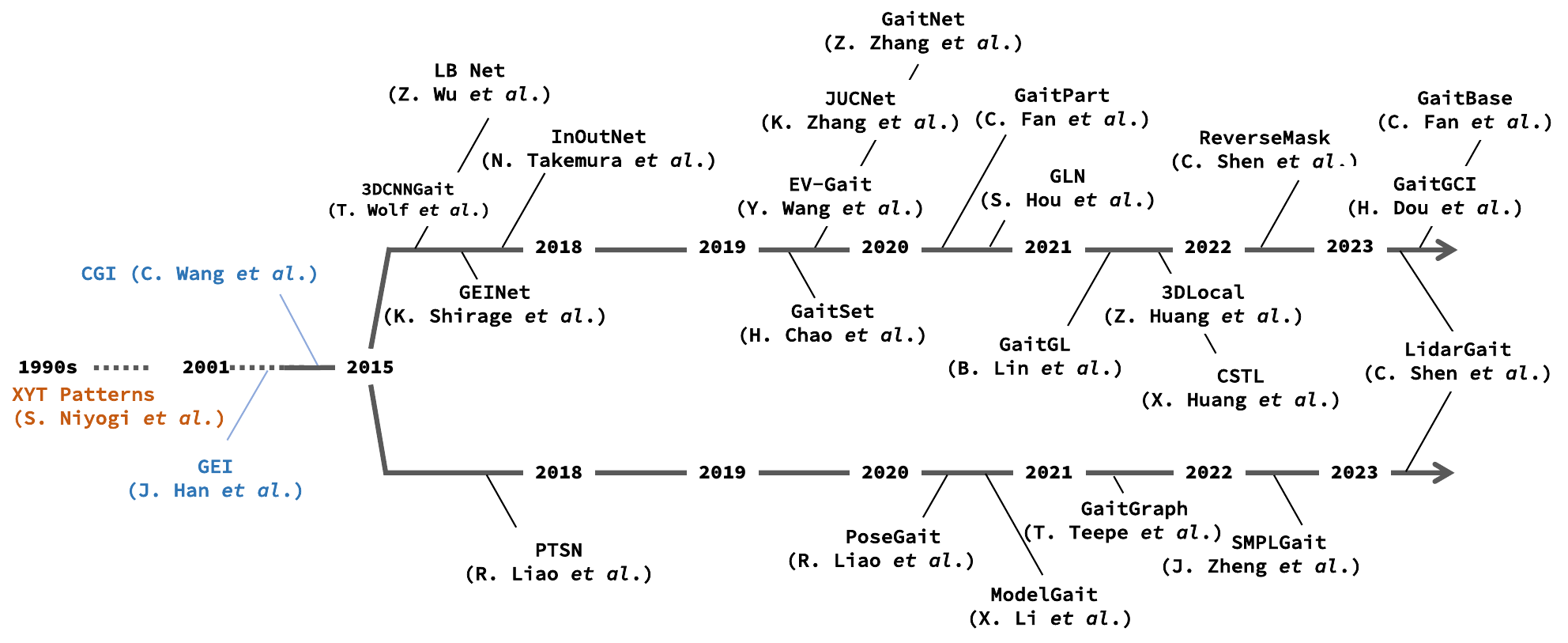}
\vspace{-5mm}
\caption{Milestone of gait recognition approaches. The reference in orange represents methods at the early stage, while references in blue indicate handcrafted feature-based methods. The references in black are representative works in the deep learning era. We categorise appearance-based methods 3DCNNGait~\cite{first3dcnngait}, GEINet~\cite{shiraga2016geinet}, LBNet~\cite{wu2016comprehensive}, InOutNet~\cite{2017inputoutput}, GaitSet~\cite{gaitsetv1,gaitsetv2}, EV-Gait~\cite{evgait}, JUCNet~\cite{cvpr2019quintuplet}, GaitNet~\cite{zhang2020gaitnet}, GaitPart~\cite{gaitpart}, GLN~\cite{gln}, GaitGL~\cite{Lin_2021_ICCV}, 3DLocal~\cite{3dlocal_2021_ICCV}, CSTL~\cite{Huang_2021_ICCV}, ReverseMask~\cite{shen2022reversemask}, GaitGCI~\cite{dou2023gaitgci}, GaitBase~\cite{fan2023opengait} on the top branch. The bottom branch presents some representative model-based methods, such as PoseGait\cite{2020liao_posegait}, PTSN\cite{PTSNliao}, ModelGait\cite{hmrgait}, GaitGraph\cite{teepe2021gaitgraph}, and SMPLGait~\cite{zheng2022gait3d}. LidarGait~\cite{shen2023lidargait} contain both model- and appearance-based features, showing conventional taxonomy is insufficient for the rapidly developing deel gait recognition.} 
\label{fig:milestone}
\vspace{-2mm}
\end{figure*}

Then the advent of the deep learning era in gait recognition powers \textit{the undergoing third stage}. This stage can be distinguished from the previous period in three main aspects: 
\textit{(i)} Deep learning techniques have revolutionized gait recognition by allowing the learning of abstract and discriminative gait features directly from input data, eliminating the need for expert knowledge and manual feature engineering. Furthermore, these deep feature-based methods~\cite{shiraga2016geinet,2016lstm_heatmap,wu2016comprehensive} consistently outperform traditional approaches that rely on hand-crafted features, such as gait energy images~\cite{han2005gei} and gait history images~\cite{GEIsurvey}. 
\textit{(ii)} Deep gait recognition continuously refreshes the scale record of pedestrians. The state-of-the-art methods can make satisfying results in datasets with hundreds of subjects. For example, the overall recognition accuracy is beyond 93\%~\cite{shen2022reversemask} under a very difficult appearance-changing setting on CASIA-B~\cite{casiab}. For evaluating over 10 thousand subjects, the cutting-edge methods can gain 98.3\%~\cite{wang2023dygait} rank-1 accuracy on OUMVLP~\cite{dataset2017oumvlp} dataset. For evaluating outdoor scenarios with over 20 thousand subjects, deep models~\cite{fan2023deepgaitv2,dou2023gaitgci} can surprisingly achieve over 70\% rank-1 accuracy on GREW~\cite{dataset2021grew}.
\textit{(iii)} With the advancement of gait recognition research in addressing challenges related to viewpoints and clothing, deep learning has opened up new possibilities for tackling even more complex problems such as in-the-wild gait recognition~\cite{shen2023lidargait,zheng2022gait3d}, multi-modal recognition~\cite{han2022licamgait}, end-to-end recognition~\cite{liang2022gaitedge}, and unsupervised recognition~\cite{fan2022gaitlu}.

\pending{Due to the rapid increase in the number of deep gait recognition methods and the great progress made in recent years, as shown in Fig.~\ref{fig:milestone}, we are motivated to develop this survey of deep learning for gait recognition. Though there are some outstanding surveys on deep gait recognition, this paper is distinguished from the existing surveys in four aspects. 
First of all, our survey is up-to-date and comprehensively covers state-of-the-art modalities, algorithms, datasets, and challenges. Meanwhile, recent surveys~\cite{2022deepsurveyACM,2021deepsurvey} have comprehensively reviewed advances using deep learning for gait recognition. Despite the fact that they provide insights into the technical aspects of deep gait recognition methods, one downside of the existing survey papers is that they do not cover remarkable achievements made in 2022 and 2023, notably by the appearance of datasets and challenges, such as gait recognition in the wild~\cite{dataset2021grew}, no label~\cite{fan2022gaitlu}, cloth-changing settings~\cite{li2023CCPG}, and 3D space~\cite{shen2023lidargait}.
Secondly, the existing surveys~\cite{dataset_survey_makihara,surveycomprehensive,2021deepsurvey,2022deepsurveyACM} follow conventional taxonomy categorizing deep gait recognition methods into the aspect of either the neural networks~\cite{2021deepsurvey,2022deepsurveyACM} (\eg CNN, RNN, GCN) or the used representations~\cite{dataset_survey_makihara,surveycomprehensive} (\ie model-based, appearance-based). However, the conventional taxonomy struggles to effectively classify new deep gait recognition methods, for example, PoseMapGait~\cite{liao2022posemapgait} and LidarGait~\cite{shen2023lidargait} utilizing both appearance- and model-based characteristics. To this end, we propose a novel taxonomy with two dimensions,~\ie deep representation and neural architectures, to provide other scholars with a systematic understanding of deep learning techniques for gait recognition.
Lastly, we provide the performance evaluations of gait recognition in four scenarios in detail,~\ie gait recognition in the cross-view setting, in the wild, in the cloth-changing setting, and in the 3D space. The comprehensive comparisons help to keep pace with the developments of diverse methods. Additionally, this paper discusses the neglected but significant topic of biometric security and privacy, addressing potential threats and challenges for exploration in the future.}

\begin{figure*}[ht!]
\centering
\includegraphics[width=0.86\textwidth]{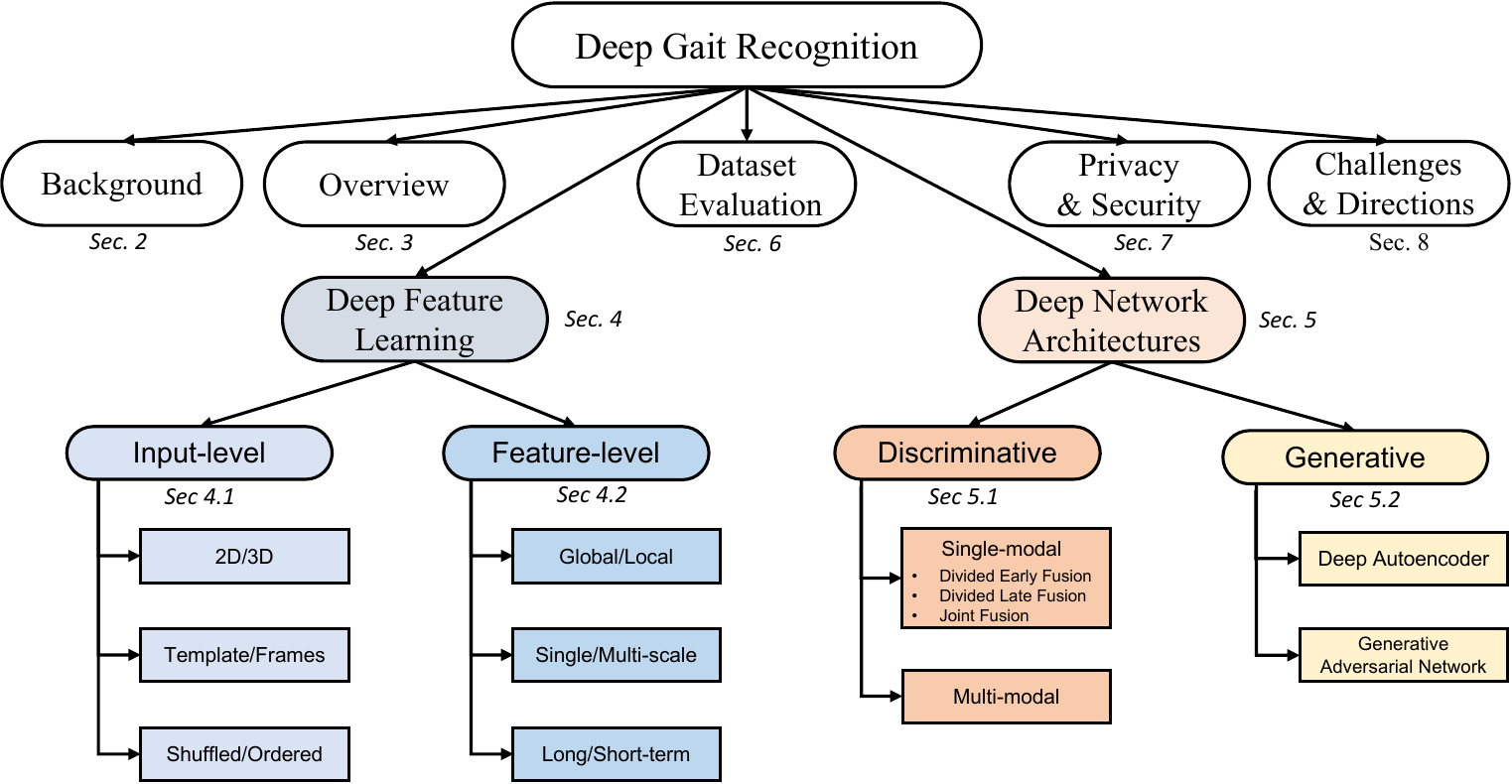}
\caption{Overall structure of this survey paper and our proposed taxonomy of the existing deep learning-based methods for gait recognition. Our proposed taxonomy, as shown in the figure, considers deep gait recognition from the perspectives of feature learning and network architecture. It provides both micro and macro scopes to enhance the understanding of deep learning-based gait recognition. Specifically, Section 4 focuses on the micro-scope of methods, exploring existing methods based on deep gait feature learning. Section 5 is presented from a more macro perspective, examining deep models architecture in gait recognition. Other fields of study also adopt a similar taxonomy by categorizing deep models into either discriminative or generative approaches.}
\label{fig:taxonomy}
\end{figure*} 

\pending{To provide an exclusive taxonomy for deep gait recognition helping to obtain a deeper understanding, this paper proposes a novel taxonomy on deep gait recognition from two perspectives: deep representations learning techniques and deep architectures designs. We explain the proposed taxonomy in detail: (1) for deep representations learning techniques in gait recognition, the deep representations learning methods can be diversely analyzed from six perspectives rather than conventional model-based/free perspectives. Specifically, we comprehensively discuss deep representations learning in gait recognition from comparisons of 2D/3D, template/frames, shuffled/ordered, global/local, single/multi-scale, and long/short-term. (2) Considering deep architectures designs, the deep architectures are divided into \textit{discriminative} and \textit{generative} models. In our humble opinion, this taxonomy provides a reasonable classification for the existing gait recognition papers, and we hope it gives readers a big picture of deep gait recognition. }

 The rest of the survey is organized as illustrated in Fig.~\ref{fig:taxonomy}. We first introduce the background of gait recognition in Section 2 and overview the main components of deep learning methods in Section 3. Section 4 reviews existing deep gait recognition methods according to the feature learning perspective. Section 5 reviews current deep gait recognition from the deep network architectures perspective. Datasets and evaluations are presented in Section 6. The security and privacy concerns are introduced in Section 7, followed by challenges and suggested directions in Section 8. The last section concludes the paper.

\section{Overview of Gait Recognition} 

\pending{\noindent \textbf{\textit{Preliminaries.}} Gait recognition is generally established into two modes: identification and verification. Verification is a one-to-one comparison used to confirm the identity, whereas identification is a one-to-many comparison used to retrieve identity from an ID gallery~\cite{jain2007biohandbook}. Unless otherwise stated, gait recognition in this survey refers to the vision-based pedestrian identification problem from sequential gait streams, which are obtained from visual sensors such as surveillance cameras, depth sensors, and Lidar sensors. }

\pending{Generally, the recognition tasks are typically categorized into two trends: \textit{open-set} and \textit{closed-set} recognition, which is called \textit{subject-independent} and \textit{subject-dependent} settings in~\cite{2021deepsurvey}. In the closed-set setting, both the training and testing phases include different samples from the same identities. However, in the open-set setting, the testing phase is tasked with recognizing a set of unseen subjects that have not been encountered yet in the training phase. The closed-set recognition has been widely studied past two decades~\cite{nixonbook}, while the more realistic and challenging open-set recognition has gained increasing interest in recent years~\cite{2022deepsurveyACM}.}

\pending{As shown in Fig.~\ref{fig:e2e} (a), a gait recognition system mainly consists of five steps:}
\begin{enumerate}
    \item  \label{item:step1} \pending{\textit{Gait Data Acquisition}: Capturing pedestrian walking sequences from various visual cameras, including RGB, infrared, and depth cameras, \textit{etc}. Most gait datasets are collected in either laboratory environments under controlled conditions~\cite{casiab,dataset2017oumvlp,shen2023lidargait}, or in-the-wild environments with complex and noisy background~\cite{zheng2022gait3d,dataset2021grew,li2023CCPG}.} 

    \item \label{item:step2} \pending{\textit{Data Annotation}: Pedestrian detection and tracking algorithms are used to extract the of-the-interest sequences. Then these sequences are labeled with detailed attributes such as identity, view, clothing condition, and carrying condition.}
    
    \item \label{item:step3} \pending{\textit{Gait Representation Generation}: To transform the raw data into gait representations such as gait silhouettes, skeletons, and optical flows. It is an essential step to maintain gait representation and to eliminate irrelevant features like the background and texture of clothes~\cite{liang2022gaitedge}.}
    
    \item \label{item:step4} \pending{\textit{Feature Learning}: Conducting a learning model to effectively represent the underlying structure of the raw gait data by a compact and abstract feature, which is a crucial step for developing a gait recognition system. While handcrafted features have been extensively studied for over two decades~\cite{GEIsurvey}, deep learning-based feature extraction has become a widely studied paradigm in recent literature~\cite{2021deepsurvey}.}
    
    \item \label{item:step5} \pending{\textit{Pedestrian Retrieval}: To retrieve the identity of a given query sequence, the trained model extracts feature representations from the query and gallery sequences and calculates their similarity. The similarity scores are used to rank the gallery sequences, resulting in a ranking list to evaluate the performance of the learning model~\cite{hou2022comprehensive}.}
\end{enumerate}

\subsection{Tasks in Gait Recognition}
By reviewing various gait recognition tasks in the literature, we briefly introduce the existing gait recognition tasks:

\noindent$\bullet$~\textit{Gait-Changing Gait Recognition.} Human gaits could be very diverse in appearance observation when observed from different viewpoints~\cite{chai2021view} or dressed in various clothes~\cite{li2023CCPG}. In addition to changes from differences in outlook, diseases like Parkinson or carrying heavy items can significantly alter the dynamic motion of human gait~\cite{applicationHC1}. Current research on gait recognition has primarily focused on appearance changes caused by different camera viewpoints~\cite{2020_view_id,chai2021view} while paying less attention to changing dynamics that can also impact gait recognition~\cite{ma2023dynamic_CVPR,wang2023dygait}. 

\noindent$\bullet$~\textit{In-the-Wild Gait Recognition.} Pedestrian retrieval can be extremely challenging in practical scenarios~\cite{shen2023lidargait,dataset2021grew,zhang2022realgait} due to unpredictable factors beyond camera viewpoints, such as variations in resolution, occlusion, frame numbers per sequence, and sequences number per subject. Albeit the research on gait recognition in the wild environment has made significant progress recently~\cite{fan2023deepgaitv2,fan2023opengait}, the existing research has been primarily focusing on gait recognition in the daytime. While investing night-time gait recognition~\cite{shen2023lidargait,song2022casiae} may be involved in cross-modality problems since the visible cameras are unfeasible at night.

\noindent$\bullet$~\textit{Unsupervised and Semi-Supervised Gait Recognition.} Unsupervised/Semi-Supervised gait recognition aims to train a model on a limited amount of labeled data~\cite{wang2022unsupervised} and utilize knowledge from a large amount of unlabeled data~\cite{fan2022gaitlu}. This task is essential for practical gait recognition since labeling large amounts of data for gait is much more time-consuming and impractical for gait than other biometrics.

\noindent$\bullet$~\textit{Heterogeneous Gait Recognition.} Heterogeneous gait recognition is the task of matching gait sequences across different modalities, such as matching a gait sequence in a surveillance video to the alternative sequence captured in an infrared camera~\cite{fu2022crossmodalpreid}. The research on cross-modality gait recognition has been neglected in the literature, while cross-modality recognition is the core of providing gait recognition at both day and night time.

\noindent$\bullet$~\textit{End-to-End Gait Recognition.} Toward learning discriminative features directly from the raw input data~\cite{Li_2021_ICCV_mvmodelgait,liang2022gaitedge,song2019gaitnet,zhang2020gaitnet}, end-to-end gait recognition extracts deep gait feature without additional steps such as silhouette segmentation or spatial alignment.

\noindent$\bullet$~\textit{Others.} In addition to the above-mentioned tasks, other important but less studied tasks in gait recognition include out-of-distribution detection~\cite{lin2022uncertaintyOOD}, which combines verification and recognition tasks to determine whether a query has a corresponding sequence in the gallery set. Another interesting area of research is anonymous gait~\cite{hirose2022anonymization}, which is relevant for privacy protection.

\begin{table*}[ht!]
\begin{center}
\caption{\pending{A comprehensive analysis of different input modalities.}}
\label{tab:datacomparision}
\vspace{-3mm}
\scalebox{0.80}{
\begin{threeparttable}
\small
\vspace{-2mm}
\begin{tabular}{lcccccc}
\midrule[1.5pt]
Data Type & Capture Devices & Resolution &  Cost & Computation  & Operational Range   & Limitations                            \\
\midrule[1.5pt]
RGB Image & Camera                  &  +++     & +       & None          & +++    & Sensitive to viewpoint and illumination            \\
Silhouette & Camera                 &  +++      & +       & ++     & +++   & Require accurate segmentation                            \\
Gait Template & Camera              &  +++     & +       & ++     & +++    & Lost micro-motion                             \\
Optical Flow & Camera               &  +++     & +       & +        & ++  & Require subject moving                            \\
Skeleton & Diverse                  &  +      & +       & ++     & ++  & Require accurate estimation                      \\
3D Human Mesh & Diverse             &  +      & +       & +++       & +   & Require accurate estimation                        \\
Depth Image & Depth camera          &  +++     & ++    & None          & +   & Unsuitable for outdoor conditions                     \\
Infrared Image & Infrared camera    &  ++   & +       & None          & +++    & Less visual details                          \\
Event Stream & Event camera         &  +++     & ++    & None          & +++    & Cannot perceive static objects                  \\
Point Cloud & LiDAR sensor          &  ++   & +++      & None          & +++    & High cost                   \\
\bottomrule\\
\end{tabular}
\vspace{-2mm}
\textit{The cost refers to the monetary cost required for the acquisition devices, while the computation denotes the computational cost of transforming the raw data into the targeted representation.}
\end{threeparttable} 
}
\end{center} \vspace{-5mm}
\end{table*}

\subsection{Various Input Data}
The environmental conditions in real-world scenarios are complex and changing, necessitating the use of various data modalities to achieve robust and reliable performance. To facilitate a better understanding of the different types of data, we have summarized their advantages and limitations in Table~\ref{tab:datacomparision}. In the following, we will provide a specific introduction to each modality.

\noindent$\bullet$~\textbf{RGB Image}.
RGB images contain a sufficient amount of information to identify different subjects. However, the texture and color biases on clothing appearance~\cite{cnnbias2018} are introduced when learning algorithms apply directly to RGB images. If RGB images are rarely taken as the input directly, the algorithm will recognize different subjects wrongly according to their appearance~\cite{li2023CCPG}. In gait recognition, clothing is regarded as a kind of variation, and gait recognition should be robust to different clothing styles. Many recent methods~\cite{zhang2020gaitnet,hmrgait,liang2022gaitedge} take RGB images as input, extracting representations robust to clothing and texture and achieving outstanding results. We believe that RGB images have great potential for gait recognition since rich information has not been fully utilized.

\noindent$\bullet$~\textbf{Silhouette}. 
A silhouette is a binary mask of a human body by removing the background and maintaining the human foreground. Human silhouettes were primarily obtained by background subtraction at an early age~\cite{piccardi2004background,casiab,dataset_casiac,soton2002}. At the same time, the advanced segmentation methods~\cite{hrnet,liang2022gaitedge,song2019gaitnet} based on deep learning can provide much better quality human silhouettes than background subtraction.
Human body silhouettes still contain an informative appearance even though color and texture are removed, but internal body structure information is partially lost in silhouettes. Besides, silhouettes are easy to be affected by clothing and camera views~\cite{liu2004qualitysilhouette}. Because of efficiency and simplicity, silhouettes were the most popular gait data in the past 20 years~\cite{jianbo2002silhouette,bblin3d,gaitsetv1,gaitpart,song2019gaitnet}. 

\noindent$\bullet$~\textbf{Gait Template}. 
Although silhouettes are efficient and simple, a sequence of silhouettes is high-dimensional. Before deep learning was widely deployed for visual recognition, it was not easy to extract features from sequential silhouettes using traditional methods like SVM~\cite{svm,begg2005SVM} and Boosting~\cite{lu2007boosting}. Han~\etal proposed Gait Energy Image (GEI)~\cite{han2005gei}, in which average cyclic silhouettes sequence into a single gait template, and such denoising processing aims to be robust for incomplete silhouettes. GEI contains comparable information to sequential silhouettes, and its data dimensionality is much less. Despite its simplicity, the GEI template is robust to many variations and achieved great success in gait recognition~\cite{GEIsurvey}. Besides GEI, some other similar features have also been proposed. They are Motion Energy Image~\cite{MEI_Template}, Gait History Image~\cite{MEI_Template}, Gait Entropy Image~\cite{GEnI}, Chrono-gait image~\cite{CGItemplate}, Gait Moment Image~\cite{GMI}, \textit{etc}. Even though deep models can handle the high data dimensionality of sequential silhouettes, some recent methods~\cite{shiraga2016geinet, 2017shiqiyuvae} still prefer GEI because of its low computational cost and robustness on noise.

\noindent$\bullet$~\textbf{Optical Flow Image}.
Gait recognition is a task to identify a subject via its walking patterns. Therefore, the motion information is significant but hardly described from silhouettes. Optical flow images can be utilized to provide more motion information than silhouettes. Castro~\etal also demonstrates that optical flow images can achieve state-of-the-art performance~\cite{2017optical}. However, the computational cost for optical flow images is relatively high, and it is also very challenging to obtain optical flow images of high quality. Recent deep learning-based FlowNet~\cite{flownet} and its successors~\cite{ilg2017flownet2,hui2018liteflownet}, can achieve relatively better optical flow images and might improve gait recognition accordingly.

\noindent$\bullet$~\textbf{Body Skeleton}.
Many methods employ body structures to extract gait motion~\cite {TraditionalModel1,TraditionalModel2,TraditionalModel3}. The gait recognition methods based on skeleton should be more robust to view and clothing variations than those based on silhouettes. However, it is not easy to extract high-accuracy human body models at the moment. Human pose estimation has achieved encouraging precision via deep learning in recent years. Those human pose estimation methods include but are not limited to DeepPose~\cite{toshev2014deeppose}, OpenPose~\cite{cao2019openpose} and HR-Net~\cite{hrnet}. Then gait recognition with human body models has returned back research of interest~\cite{PTSNliao,teepe2021gaitgraph,gpgait}, and many datasets~\cite{shen2023lidargait,2020anPoseDataset,dataset2021grew} with pose annotations presented to advance model-based gait recognition.

\noindent$\bullet$~\textbf{Human Mesh}. 
Mesh is a type of 3D representation that consists of a collection of vertices and polygons to define the exact shape of an object~\cite{wang2018pixel2mesh}. Compared to skeletons, the human mesh can provide more structural information. There are various human mesh recovery methods~\cite{meshrecovery1,meshrecovery2} to construct a complete 3D body model. ModelGait~\cite{hmrgait} fine-tunes a mesh recovery model on a gait dataset and distinguishes different subjects via extracted structural parameters, showing the promising performance of utilizing human mesh as auxiliary supervision information. Gait recognition based on body meshes will be an exciting topic in the future with the improvement of human body mesh estimation accuracy.

\noindent$\bullet$~\textbf{Depth Image}.
Unlike color images, depth images can provide a 3D structure of bodies since each pixel value is the distance between the object and the camera. The low-priced depth cameras like Kinect~\cite{kinectsensor} provide the possibility for gait recognition using depth images. In~\cite{dataset_tum_gaid}, traditional GEI is compared with depth-based templates such as Depth-GEI, DGHEI, and GEV, and experiments show that depth templates can achieve better performance. A comprehensive review on gait recognition with depth images can be found in~\cite{depthgaitreview}, introducing public depth datasets and most methods with depth images. Depth image-based gait recognition has a primary challenge in that a depth camera can only capture data in a range of 10 meters. Besides, the active infrared light from depth cameras will decrease dramatically with the distance and can also be disturbed by sunlight. For those reasons, gait recognition with depth images is difficult to deploy into an outdoor system to capture gait from a distance.

\noindent$\bullet$~\textbf{Dynamic Event Stream}.
Event stream cameras can capture high-speed movements without blurs. The dynamic vision sensors can capture microsecond-level pixel intensity changes as events by a class of neuromorphic devices. By converting the event stream into image-like representations~\cite{evgait}, CNN-based methods can contribute to extracting discriminative features from this data modality. Event streams may provide much more promising performance from their ability to capture dynamic fine-grained motion. In the literature, EV-Gait~\cite{evgait} is the first work on dynamic vision sensors for gait recognition. It achieved nearly 96\% recognition accuracy in a real-world setting and comparable performance with state-of-the-art RGB-based gait recognition methods on the CASIA-B benchmark. However, more studies on this new sensor are needed, and it is great potential for event cameras to deploy gait recognition systems in the future. 

\pending{\noindent$\bullet$~\textbf{Point Clouds}. Point clouds are typically produced by IoF sensors like LiDAR, which is capable of facilitating outdoor gait recognition~\cite{shen2023lidargait,ahn20222vgait,han2022licamgait} with precise 3D information. LiDAR sensor provides not only robust gait representation in many challenging scenarios such as poor illumination and occlusion, but it can also perform gait recognition at a large range of distances. Point clouds have been preferred in recent years because they provide precise 3D geometry. LidarGait~\cite{shen2023lidargait} demonstrates that LiDAR-based gait recognition can outperform traditional camera-based methods by a large margin in challenging conditions like poor illumination and cross-view scenarios. Besides, point-based gait recognition shall be in favor of privacy-preserving scenarios such as nursing homes, and it is potentially superior to camera-based methods in biometrics protection with less sensitive information. }

\subsection{Feature Learning for Gait Recognition}
The gait feature learning methods can be generally categorized into handcrafted feature-based methods (Sec.~\ref{ref:sub-handcraft}) and deep feature-based methods (Sec.~\ref{ref:sub-deepfeature}).

\subsubsection{Handcrafted feature-based gait recognition}
\label{ref:sub-handcraft}

Gait recognition methods can be broadly categorized into two groups: model-based and appearance-based methods, depending on whether they explicitly model the structure of the human body~\cite{nixonbook}.

Model-based methods can be further categorized into structural and motion models. \textbf{Structural models}~\cite{structure1,structure2} employ static body parameters such as stride length and cadence as clues. For example, Boulgouris~\etal~\cite{structure4} utilized labeled structural gait silhouettes to recognize different subjects by extracting discriminative component-wise representations. On the other hand, \textbf{motion models} focus on dynamic motion features. Early attempts include using phase-weighted magnitude spectra~\cite{cunado1997phase} and Fourier description from the motion of the hip and thigh~\cite{nixon1996automatic}. Some works combine both static body structure and dynamic motion to improve accuracy~\cite{cunado1999gait,hmrgait}.

The \textbf{appearance-based} methods directly extract gait-relevant features from the input without constructing human models. These methods dominate gait recognition due to their effectiveness and efficiency. For instance, Shi~\etal~\cite{shijianbosilhouette} proposed a viewpoint-dependent silhouette-based baseline method in 2002, while Wang~\etal~\cite{dataset_wang2003silhouette} applied principal component analysis (PCA) to reduce feature dimensionality. Many effective gait templates~\cite{GEnI,han2005gei,GMI} further led to the migration of gait recognition methods towards using gait templates as features, even in the deep learning era.

\subsubsection{Deep feature-based gait recognition}
\label{ref:sub-deepfeature}
\begin{figure}[htbp]
\centering
\includegraphics[width=0.4\textwidth]{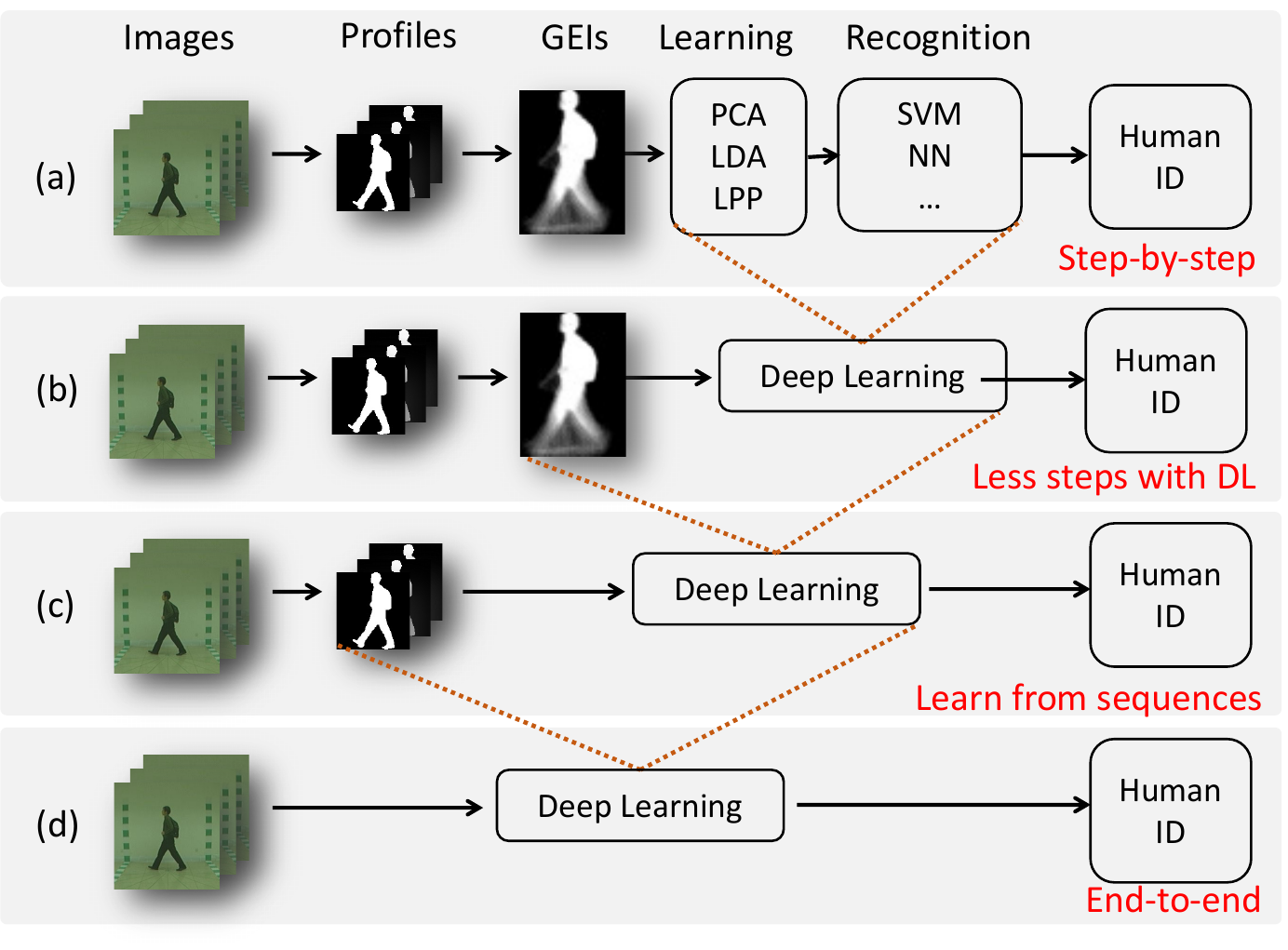}
\caption{Four typical workflows on deep gait recognition. Figure adapted by~\cite{song2019gaitnet}.}
\label{fig:e2e}
\end{figure}

Deep feature learning has emerged as the dominant approach for gait recognition, revolutionizing how representations are extracted. While traditional handcrafted feature-based methods rely on prior knowledge and expertise to design descriptors, deep feature learning methods focus on network architecture and loss functions. With deep learning, neural networks automatically extract gait features through a series of stacked layers. As shown in Fig.~\ref{fig:e2e}, early methods in this area overlooked temporal information within gait sequences and heavily relied on gait templates~\cite{CGItemplate,han2005gei}. However, recent works~\cite{song2019gaitnet,2019zhangACL} have made significant strides in extracting frame-level features directly from a sequence of silhouettes. The typical pipeline of deep gait recognition involves front-ground segmentation, gait alignment, and feature extraction. Lastly, some recent works~\cite{liang2022gaitedge,hmrgait} have explored all-in-one models that learn all necessary steps from input to output.

\begin{figure*}[htbp]
\centering
\includegraphics[width=0.75\textwidth]{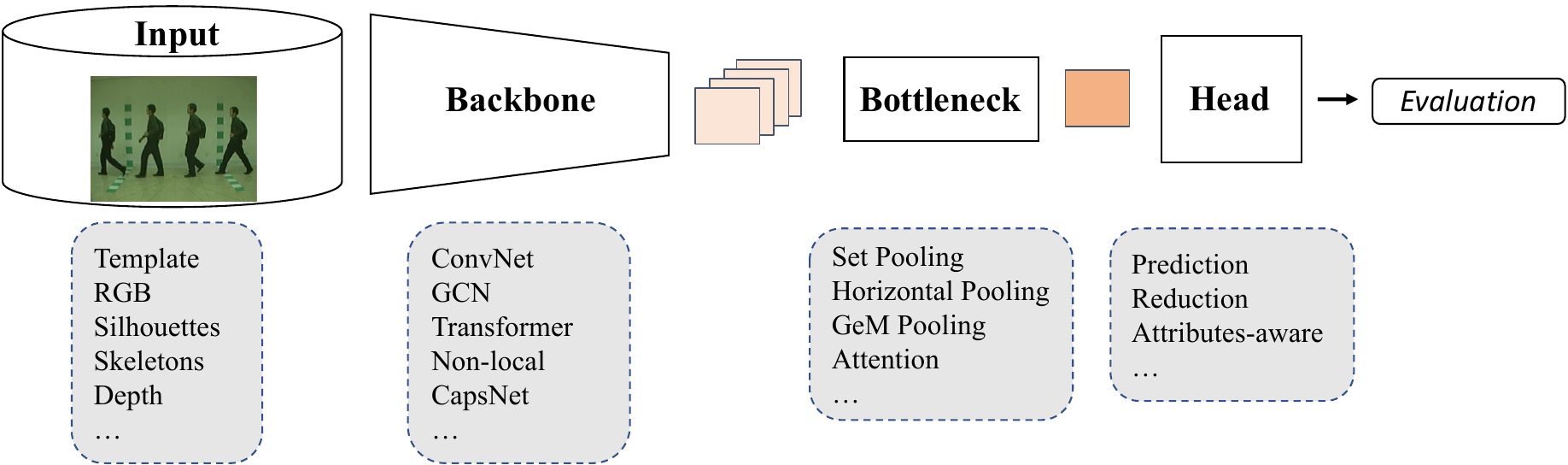}
\caption{The typical workflow of deep gait recognition models.}
\label{fig:backbone}
\end{figure*}

\section{A Glimpse of Deep Gait Recognition}
Deep gait recognition methods generally comprise three main components: a backbone for feature extraction, a bottleneck for spatial-temporal feature aggregation, and a head for representation mapping.

\subsection{Backbone Networks}   
The backbone of a deep gait recognition model serves as the feature extractor, transforming input data into deep abstract representations. Traditional gait recognition methods used hand-crafted features like gait templates~\cite{han2005gei,GEnI,2017trajectory}. However, with the emergence of deep learning, learned features from deep neural networks have shown significant improvements over manually designed features.

Convolutional Neural Networks (CNNs) are the seminal backbone networks in deep gait recognition models. While early works used 2D CNNs and achieved better results than handcrafted features~\cite{cnn,2015TMM,zhang2020gaitnet}, they lacked effectiveness in capturing temporal information. To address this, more advanced backbones~\cite{first3dcnngait,Lin_2021_ICCV} with 3D convolutional layers were introduced to extract robust spatiotemporal features for improved identification.

Recent literature has explored gait-specific backbone networks to enhance performance further. For example, focal convolutions~\cite{gaitpart,Lin_2021_ICCV} were proposed to capture fine-grained gait patterns from horizontally partitioned inputs, and 3D local convolutions aimed to obtain component-specific information from the pedestrian body adaptively~\cite{3dlocal_2021_ICCV}. Additionally, graph convolution networks have been adopted for models taking skeletons as input, effectively extracting structural and dynamic information~\cite{teepe2021gaitgraph}. 

Notably, a significant difference between gait recognition and other visual recognition tasks lies in the use of relatively shallow backbone networks (ranging from 4 to 8 layers). Many visual recognition tasks exploit very deep models with tens or hundreds of layers. We analyze this phenomenon mainly because gait recognition often utilizes silhouettes and skeletons, which contain lower information entropy compared to RGB images or videos. This observation highlights the importance of using the original RGB modality to take advantage of deeper models for improved precision.

\subsection{Bottleneck Networks} 
Bottleneck networks in gait recognition are built for aggregating dynamic gait information through temporal modeling and enhancing the discriminative features via spatial manipulation.

Early approaches often overlooked the importance of designing bottleneck networks, relying on fully connected layers for feature reduction and global feature learning. However, these methods were sensitive to noise and could overfit on poorly segmented data, leading to significant performance degradation~\cite{2015TMM,shiraga2016geinet,wu2016comprehensive}. Therefore, Horizontal Pyramid Pooling~\cite{gaitsetv2}, Patch Pyramid Mapping~\cite{zhang2022realgait}, and Generalized-mean Pooling~\cite{Lin_2021_ICCV} were proposed to capture fine-grained gait cues, which leveraged partial features to prevent overfitting. Hou~\etal~\cite{gln} introduced feature lateral learning, where the inherent feature pyramid utilizes multiple-scale features to enhance gait representations. Besides, Huang~\etal~\cite{Huang_2021_ICCV} measured the importance of different parts across frames, which exploited the most discriminative parts and generated more robust spatio-temporal representations. Considering temporal representations modeling, Set Pooling~\cite{gaitsetv2} proposed various feature pooling strategies along the time dimension. The Micro-motion Capture Module~\cite{gaitpart} and Adaptive Temporal Aggregation~\cite{Huang_2021_ICCV} made use of attention mechanism to extract gait patterns in the long short-term manner, and recurrent neural networks~\cite{zhang2019gait} are also able to perform adaptive temporal representation from sequential inputs.

\subsection{Head Networks} 
In gait recognition, head networks following the bottleneck are optional but serve specific purposes. In object detection, deep models typically have two heads for object recognition and bounding box localization~\cite{yolo}. Similarly, gait recognition can employ various head networks for multi-task gait recognition.

The verification head~\cite{wu2016comprehensive} and identification head~\cite{gln} are the most commonly used heads in deep gait models. While identification heads achieve satisfying precision with a strong backbone and bottleneck, it was noted that their feature dimensionality may not be compact enough for practical applications. To address this, the compact block~\cite{gln,3dlocal_2021_ICCV} was introduced to reduce representation dimensionality and memory usage. Moreover, other head networks have been proposed, such as quality-aware~\cite{hou_2022_quality}, view-aware~\cite{chai2021view}, condition-aware~\cite{wu2020condition}, and gender-aware~\cite{dataset2021versatilegait}, offering contributions for multi-task gait recognition.

\subsection{Loss Functions}
\label{sec:metriclearning}
The loss functions are designed to measure the similarity between samples in the embedding space. By minimizing the loss function during training, the deep networks learn to map similar samples to nearby points and dissimilar samples to faraway points in the embedding space. Gait recognition conducts deep metric learning using many commonly used losses, including cross-entropy, contrastive, and triplet losses.

\textbf{Cross-entropy loss}. 
The cross-entropy loss~\cite{LSoftMax} is a classical loss function, which is widely used in many classification tasks. By measuring the difference between the predicted probability distribution and the actual probability distribution of the target class, cross-entropy loss is to minimize the difference of the distribution, making the probability distribution of the model output as close as possible to the ground-truth, the identification labels. The loss function is
$$L_{ce} = -\frac{1}{P\times K}\sum_{i=1}^{P\times K}\sum_{n=1}^{N}q_n^ilog(p_{n}^i),$$
Where $N$ is the number of all identities in the training set, $P\times K$ denotes the number of samples in a mini-batch, $p_n$ denotes the probabilities of a sample $i$ belonging to identity $n$, and $q_n$ is a binary indicator (0 or 1), if $n=y$ then $q_n=1$ (taking the y-th identity as an example). However, the cross-entropy loss may not be the optimal choice for gait recognition due to several reasons. Firstly, the number of classes can be very large in gait recognition. For instance, the OUMVLP dataset~\cite{dataset2017oumvlp} contains over 10K subjects, and GREW has over 20K subjects. Secondly, the inter-class variation among different classes can be significant, making it difficult for the cross-entropy loss to capture the similarity/dissimilarity between samples effectively.

\textbf{Contrastive loss}. 
Given the challenge of handling a large number of classes, pairwise losses have emerged as an alternative to cross-entropy loss in gait recognition. A commonly used pairwise loss is the contrastive loss, which enforces a margin between the distances of positive pairs and negative pairs. Specifically, it encourages the distance between positive pairs to be less than a margin value, while ensuring that the distance between negative pairs is greater than this margin value.
$$L_{con} = \frac{1}{2N}\sum_{n=1}^{N}(y_nd_n^2+(1-y_n)max(m-d_n,0)^2),$$
where $N$ is the number of training sample pairs and $d_n$ is the dissimilarity score(usually calculated by L2 norm distance or inner product) of the $n$-th pair. $y_n$ is a binary indicator (0 or 1), setting to one when the samples in the $n$-th pair from the same identity. The hyper-parameter $m$ in contrastive loss refers to the margin.

\textbf{Triplet loss}. 
Triplet loss~\cite{tripletloss} is widely used in recent state-of-the-art methods~\cite{fan2023opengait}. Instead of using pairs, this loss takes distance triplets $(anchor,\; positive,\; negative)$ as input. The loss pulls the positive samples close to the anchor and pushes the negative away from the anchor. In order to prevent the features from converging into a small space, the distance between the anchor-negative pair should be at least one margin $m$ farther than that of the anchor-positive pair. 
$$L_{tri} = \frac{1}{N_{tp+}}\sum_{\substack{a,p,n\\y_a=y_p\neq y_n}}max(m+d(a,p)-d(a,n),0),$$
where $N_{tp+}$ denotes the number of triplets of non-zero loss terms in a mini-batch, $a, p, n$ stand for anchor, positive and negative, respectively. $d(a,p)$ and $d(a,n)$ denotes the distance between anchor-positive and anchor-negative respectively. The hyper-parameter $m$ in contrastive loss refers to the margin.

\textbf{Other losses}. 
The aforementioned loss functions achieve significant performance for gait recognition, but these losses are also widely used in many tasks like image classification, face recognition, and person re-identification. In the context of human-centre tasks, many challenges such as cross-view and changing appearance have not been well solved. Toward solving challenges of the human-centric gait recognition, recent research has designed gait-specific loss functions including, Angle Center Loss~\cite{2019zhangACL}, Quintuplet Loss~\cite{cvpr2019quintuplet}, and View Loss~\cite{chai2021view,Chai_2022_CVPR}.

\subsection{Evaluation}
Evaluating the performance of gait recognition algorithms depends on two primary measurements: evaluation \textit{metrics} and \textit{protocols}.

\pending{In gait recognition, the choice of evaluation metrics depends on the specific recognition modes being considered. In the one-to-one verification mode, the performance is typically evaluated using operating characteristic curves (ROC), which provide a visual representation of the trade-off between the true positive rate and the false positive rate. On the other hand, the one-to-many identification mode employs several primary metrics, including Cumulative Matching Characteristics (CMC) curves~\cite{casiab,wu2016comprehensive}, mean Average Precision (mAP)~\cite{hou2022comprehensive,li2023CCPG}, and mean Inverse Negative Penalty (mINP)~\cite{zheng2022gait3d,zheng2022gaitmultihop}. }

\pending{Evaluation protocols may vary across different datasets, as each dataset often emphasizes specific aspects or challenges of gait recognition. A dataset may include multiple evaluation settings, resulting in multiple evaluation protocols. Since a survey by Hou~\etal has provided a comprehensive study on evaluation protocols for gait recognition, we recommend referring to~\cite{hou2022comprehensive} for more detailed information and a more thorough understanding.}

\section{Deep Representations Learning} 
In deep gait recognition, representation learning involves the extraction of abstract deep gait descriptors using deep networks. In this section, we discuss deep gait feature learning from \pending{input-level (Sec.~\ref{sec:inputlevel})} and \pending{feature-level (Sec.~\ref{sec:featurelevel})} perspectives.

\subsection{From Input-level Perspective}
\label{sec:inputlevel}
Deep models learn diverse gait representations based on the type of input data they receive. Depending on the characteristics of input data, such as ordered frames or shuffled frames, deep models can capture different aspects of gait information. For instance, when ordered frames are used, deep models can effectively capture dynamic motion, while micro-motion features may be neglected when shuffled frames are used. In this section, we explore deep representation learning from an input-level perspective, focusing on the following three aspects.

\subsubsection{2D/3D representation learning}
\textit{2D representation learning} extracts geometric information from data captured by 2D visual sensors~\cite{2022deepsurveyACM,nixon1996automatic,nixonbook}. With the simplicity and efficiency of the 2D representation, 2D representations have dominated gait recognition for over 30 years. The commonly used 2D representations range from images, silhouettes, skeletons, and optical flows. Among them, silhouettes are the most popular 2D representation because silhouettes are easy to get and have precise body shape information. Using silhouettes as gait representation, many work~\cite{song2019gaitnet,dou2022metagait,fan2023opengait} achieve satisfying performances in both indoor and outdoor environments. However, silhouettes lack representing inter-frame motion information, which inspires methods using optical flows to exploit spatial-temporal cues explicitly~\cite{2017optical}. Besides, model-based methods also utilize 2D skeletons~\cite{2020liao_posegait,teepe2021gaitgraph} as input, trying to learn better view-invariant features, but model-based methods suffer from the precision of the pose estimation model~\cite{dataset2021grew}. Therefore, more research~\cite{zhang2020gaitnet,hmrgait,liang2022gaitedge} focuses on disentangling gait-unrelated and gait-related features from RGB images.

\textit{3D representation learning} aims to learn features from 3D data captured from either 2D-to-3D estimation models or advanced sensors. 3D representation is preferred because of its outstanding performance in handling viewpoint changes compared to 2D representation~\cite{shen2023lidargait}. Estimation-based 3D representation, such as 3D skeletons~\cite{2020liao_posegait,peng2023multimodal,cao2019openpose} and 3D human mesh~\cite{zheng2022gait3d,meshrecovery1,meshrecovery2}, can enhance the performance of gait recognition. However, estimation-based 3D representation is sensitive to many factors such as illumination and resolution, which limit its performance in challenging real-world scenarios. To address this issue, more recent research has focused on capturing depth images of human bodies~\cite{depthgaitreview, dataset_tum_gaid, dataset_iits_depthGait} from depth cameras. However, depth cameras are only feasible indoors and cannot be used when the distance to pedestrians is over 10 meters. In recent years, Shen~\etal~\cite{shen2023lidargait} have used LiDAR sensors to obtain precise 3D structures of the human body and demonstrated LiDAR-based gait recognition with robust cross-view performance, outperforming 2D representation-based methods by a large margin.

\subsubsection{Template/frames-based representation learning}
Representation learning can be categorized into \textit{frames-based} methods and \textit{template-based} methods, according to whether utilizing a whole sequence of frames as input or not.

\textit{Template-based} methods receive a gait template as input for gait feature extraction. These methods were popular before the deep learning era when traditional learning methods like SVM~\cite{svm} and Boosting~\cite{lu2007boosting} struggled to extract effective features from high-dimensional data. Therefore, researchers designed various gait templates~\cite{han2005gei,CGItemplate,GEnI,FDEI,MEI_Template}, and these templates achieved high accuracy in many datasets~\cite{surveyGEI}. For example,  Gait Energy Image (GEI) represents the average of cyclic silhouettes, preserving static information while partially losing gait motion information. As the most popular gait template, GEI~\cite{han2005gei} is widely used all the time because it is computationally efficient and temporally robust. However, with the advancement of deep learning, many recent methods~\cite{song2019gaitnet,2019_joint_zhangyuqi,Huang_2021_ICCV,bblin3d,zheng2022gaitmultihop} have focused on frames-based representation learning to extract dynamic motion features for better performance.

\textit{Frames-based} methods utilize all frames of gait sequences as input, enabling the extraction of fine-grained spatial information from diverse modalities such as silhouettes~\cite{song2019gaitnet,2019_joint_zhangyuqi,Huang_2021_ICCV}, body skeletons~\cite{teepe2021gaitgraph}, SMPL~\cite{hmrgait,zheng2022gait3d}, and 3D point clouds~\cite{han2022licamgait,shen2023lidargait,ahn20222vgait}. These methods also leverage sequential inputs to model gait motions using various temporal modeling techniques, including set-based~\cite{gaitsetv1}, long short-term~\cite{bblin3d}, shift-based~\cite{zheng2022gaitmultihop}, and attentive modeling~\cite{Huang_2021_ICCV}. By combining spatial and temporal information, frames-based methods enhance recognition robustness and accuracy, making them a prominent trend in deep gait recognition.

\subsubsection{Shuffled/ordered representation learning}
\label{sequence}

\textit{Shuffled learning} utilizes unordered sequences as model input to aggregate set-based temporal information, making it robust to frame permutations and extendable to cross-scenes scenarios.  This approach has been widely used in deep gait recognition~\cite{2015TMM,gaitsetv1}. By regarding gait as a set of gait silhouettes, set-based shuffled learning is robust to scenarios when the input samples contain discontinuous frames or have a frame rate different from the training dataset~\cite{gaitsetv2}. Although ordered representation learning has shown outstanding performance in in-the-lab datasets, models using shuffled inputs outperform those using ordered inputs in outdoor datasets~\cite{zheng2022gait3d, dataset2021grew,shen2023lidargait}. This is because people walk at varying speeds and routes in the real world, unlike the in-the-lab dataset setting. However, recent research~\cite{ma2023dynamic_CVPR,fan2023deepgaitv2} has demonstrated that incorporating residual learning and increasing the number of layers can improve recognition accuracy on in-the-wild datasets, even when using ordered inputs for temporal feature extraction.

\textit{Ordered} learning involves taking sequences in their natural order for motion modeling. For tasks that are based on sequences in computer vision, using ordered frames is the most straightforward and natural way to model fine-grained motion~\cite{twostream}. Ordered representation learning models temporal features via Recurrent Neural Networks (RNNs)~\cite{2020_RNN_skeleton}, Convolution Neural Networks (CNNs)~\cite{cnn}, Long Short-term Memory Networks (LSTM)~\cite{2016lstm_heatmap}, or Transformer Networks~\cite{fan2023deepgaitv2}. This ordered motion learning is effective for in-the-lab datasets that contain gait sequences only in walking status~\cite{gaitsetv1,gaitpart,Lin_2021_ICCV}. However, the use of temporal learning from ordered inputs has underperformed when applied to in-the-wild datasets~\cite{zheng2022gait3d,dataset2021grew}. This is due to two possible reasons, as shown by the results reported in GREW~\cite{dataset2021grew}, Gait3D~\cite{zheng2022gait3d}, and SUSTech1K~\cite{shen2023lidargait}. Firstly, people may walk at diverse speeds and routes in real-world scenes~\cite{song2022casiae}. Secondly, the quality of gait representations in the inputs might be low, disrupting dynamic motion learning. To address this, recent studies have introduced deep residual learning~\cite{fan2023deepgaitv2} and continuous frame sampling~\cite{zheng2022gaitmultihop} to capture gait motion patterns from ordered inputs in the real-world scenario.

\subsection{From Feature-level Perspective}
\label{sec:featurelevel}
In this part, we discuss deep representation learning from a feature-level perspective.

\subsubsection{Global/local representation learning}

\textit{Global representation learning} refers to extracting holistic features from the entire human body. Early gait recognition methods based on deep learning used global representation learning. For example, Hossain~\etal~\cite{firstdlgait} introduced deep learning into gait recognition by using several layers of convolution to extract deep gait features from the GEI templates. Another example is that Wu~\etal~\cite{wu2016comprehensive} employed a 3-layer CNN to process GEIs of two gait sequences from two branches, utilizing verification mode to determine whether the two samples belong to the same person. Moreover, global representation can also be learned from human skeletons estimated from images by pose estimation models such as OpenPose~\cite{cao2019openpose}, or from 3D point clouds~\cite{ahn20222vgait} gained by LiDAR sensors. These model-based feature learning methods, as demonstrated in studies~\cite{PTSNliao,20183dpose,2019TGLSTM}, have shown robust performance in cross-view recognition. However, gait recognition is an instance-level recognition task that relies on subtle differences to distinguish people. Therefore global representation is suboptimal to methods that explicitly learn features from multiple local regions.

\textit{Local representation learning} focuses on capturing fine-grained features by exploiting spatial representations from specific regions of human bodies. In the traditional era, Liu~\etal~\cite{usflabel} highlighted the significance of different body parts having distinct shapes and moving patterns. Recent works have proposed effective methods for learning local representations. For example, Fan~\cite{gaitpart} introduced GaitPart, utilizing focal convolutions on horizontally partitioned body parts. Shen~\etal~\cite{shen2022reversemask} addressed the limitation of hard partitioning and proposed ReverseMask, randomly generating masks to force CNNs to learn local features. Horizontally Pyramid Pooling, employed in methods like~\cite{gaitpart, gaitsetv1}, has become a widely used technique for separating holistic features into multiple horizontal strips to capture local patterns. Local features can also be extracted from pose-driven regions of interest (RoI)\cite{posereid}, attention regions of appearance\cite{attentionReid}, body components through parsing~\cite{usflabel}, or patch-level approaches~\cite{patchJapan}. Research~\cite{Lin_2021_ICCV,2019zhangACL} has demonstrated that local representation learning-based methods can achieve better results.

In conclusion, global features focus more on holistic information, whereas local features focus on partial information. Global features contain more coarse information, while local features contain fine-grained information. GaitGL~\cite{Lin_2021_ICCV} combines the two kinds of features for better performances, which is a reasonable solution for learning better gait representations.

\subsubsection{Single/multi-scale representation learning}
\textit{Single-scale representation learning} involves the process of learning representations at a fixed scale, where a single level of abstraction is used to represent the features of the input data. This method has been extensively employed in traditional machine learning algorithms such as XYT template~\cite{niyogi1994analyzing} and gait in eigenspace~\cite{murase1996movingsilhouttes}. Additionally, single-scale representation learning has been widely used in various visual tasks in the deep learning era due to its implementation simplicity and low computational cost. For instance, GEINet~\cite{shiraga2016geinet} takes GEIs at a holistic scale as input and extracts deep features for recognition, like other single-scale learning methods~\cite{2015TMM,2017inputoutput,wu2016comprehensive}.

\textit{Multi-scale representation learning}involves learning representations from multiple levels to capture different scales of features. This approach has been widely adopted in object detection and person re-identification~\cite{HPPreid,mao2018multi} and is effective for gait recognition because it enhances the discriminative capabilities of different human body parts.
One example of this is the use of horizontal pyramid pooling~\cite{HPPreid} in GaitSet~\cite{gaitsetv2,gaitsetv1}, inspired by multi-scale learning in person re-identification. This method has been employed in many subsequent works, such as GaitGL~\cite{Lin_2021_ICCV}, GaitPart~\cite{gaitpart}, and CSTL~\cite{Huang_2021_ICCV}. Additionally, GLN~\cite{gln} merges the features from different stages in a top-down manner to enhance the gait representations with multi-scale robustness.

\subsubsection{Long/short-term representation learning}
\textit{Long-term representation learning} extracts human dynamic motion at a temporally large scale. At the input level, template-based methods use a sequential set of gait images to construct a gait template~\cite{GEIsurvey}. At feature-level long-term representation learning, long-term features can be captured by temporal pooling~\cite{gaitsetv1,gaitsetv2}, temporal attention~\cite{Huang_2021_ICCV}, or LSTM~\cite{2016bmvclstm}. While GaitSet~\cite{gaitsetv1} and GLN~\cite{gln} focus on global set-level temporal feature extraction, but they neglect inter-frame dependency modeling. To address this, GaitGL~\cite{Lin_2021_ICCV} employed multiple stacked layers of 3D convolutions to model inter-frame features. Differently, CSTL~\cite{Huang_2021_ICCV} proposes a feature selective pooling module that preserves the most discriminative spatial local features into a final representation. As experiments presented in outdoor datasets~\cite{dataset2021grew,shen2023lidargait,zheng2022gait3d}, long-term temporal cues perform robustly in complex scenarios where human walks in diverse routes and cameras capture gait sequence from different viewpoints.

\textit{Short-term representation learning} involves extracting gait micro-motion within the adjacent frames. Current literature proposed various strategies for modeling gait local patterns, including LSTMs~\cite{2016lstm_heatmap}, 1D convolutions~\cite{gaitpart,Huang_2021_ICCV}, 3D convolutions~\cite{first3dcnngait}, and temporal shift~\cite{zheng2022gaitmultihop}. LSTMs, a type of recursive neural network, are commonly used in the field of speech recognition and handwriting recognition for sequential signals and have also been widely used in gait recognition, such as PTSN~\cite{PTSNliao}, and Zhang's GaitNet~\cite{zhang2019gait}. GaitPart~\cite{gaitpart} extracted local temporal clues by 1D convolutions and subsequently aggregated them into compact features with inter-frame dependency. Recent methods~\cite{Lin_2021_ICCV,Chai_2022_CVPR} using 3D convolutions extract spatial-temporal features simultaneously, enhancing the ability to capture fine-grained gait dynamics with discriminativeness. However, while short-term representation learning has achieved outstanding results on laboratory datasets, these in-the-lab datasets capture gait sequences with humans walking continuously in limited viewpoints. When conducted on in-the-wild datasets, methods with temporal convolutions to extract short-term gait cues underperform methods based on long-term representations~\cite{shen2023lidargait,zheng2022gait3d}.

\begin{figure}[htbp]
\centering
\includegraphics[width=0.35\textwidth]{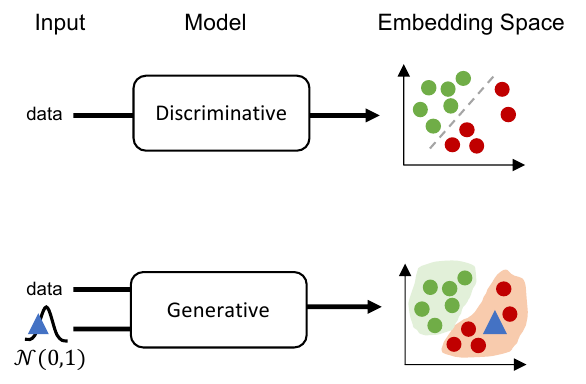}
\caption{Illustration of the discriminative and generative model.}
\label{fig:disandgen}
\end{figure} 

\section{Deep Models for Gait Recognition} 
\label{sec:architecture}
\pending{In this section, we categorize deep gait models into discriminative and generative models. As illustrated in Fig.~\ref{fig:disandgen}, discriminative models are trained to learn the classification boundary between different identities, while generative models do not learn the decision boundary but model the underlying distribution of the data, enabling the generation of new samples from the given data.}

\subsection{Discriminative Model}
The discriminative models are the methods of directly learning decision boundaries to recognize human gait. To learn discriminative gait representations, these models can be divided into two categories, including the single-modal model and the multi-modal model. 

\begin{figure}[htbp]
\centering
\includegraphics[width=0.45\textwidth]{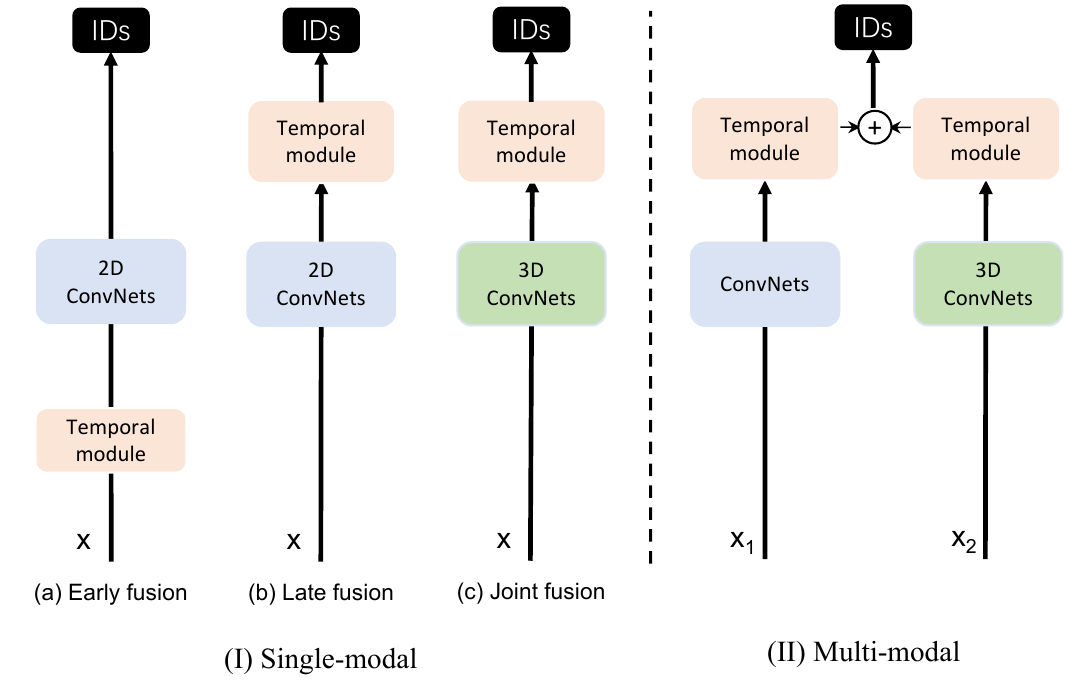}
\vspace{-2mm}
\caption{Illustration of four types of models in discriminative architecture.}
\label{fig:architecture}
\end{figure} 
\subsubsection{Single-modal architecture}
Learning deep representation from a single modality is a classical and most commonly used method in many visual recognition tasks. The single-modal models dominate the study of deep gait recognition because these methods are often more computationally efficient than multi-modal models. As gait recognition involves spatiotemporal modeling to preserve appearance and motion features, we summarize single-modal models on deep gait recognition into three classes according to the position where spatiotemporal information is fused together in the deep networks. These three classes of single-modal models are, namely, (1) divided early fusion models, referring to models in which spatial information extraction is followed by temporal modeling. (2) divided late fusion models, capturing spatial information and then aggregating temporal motion. (3) joint fusion models, which consider spatiotemporal modeling simultaneously. The single-modal gait recognition methods also follow the same development process, as detailed in the three following aspects.

\vspace{3mm} \noindent{\textbf{{Divided Early Fusion (T+S) Model}}}

\pending{\vspace{1mm} Divided early fusion models refer to the category of methods that fuse gait inputs at the very initial stage and then extract gait representation with spatiotemporal information as illustrated in Fig.~\ref{fig:architecture}(a). This architecture is also called template-based architecture in other literature. As many works~\cite{shiraga2016geinet,wu2016comprehensive,song2019gaitnet} indicate that divided early fusion architecture obstructs deep feature extraction spatially and temporally, this architecture had been adopted in many early methods and has still been utilized in many recent works~\cite{2020ffgei} because of its computational friendly.}

\pending{Among various template-based methods, gait energy image~\cite{han2005gei} is the one that cannot be neglected from the history of gait recognition. By averaging gait silhouettes along the temporal dimension, GEI showed its simplicity and efficiency in cross-view datasets with over 10,000 subjects~\cite{dataset2017oumvlp}, making it become the most dominant template among traditional gait recognition methods. Therefore GEI continued to be the first choice of data modality to learn gait features when deep learning first meets gait recognition. Particularly, Shiraga~\cite{shiraga2016geinet} proposed a CNN-based model, named GEINet, which composes only two sequential triplets of convolution, pooling, and normalization layers, along with two subsequent fully connected layers. GEINet utilized CNN on GEI, and marginally outperformed the first deep gait recognition method paper~\cite{firstdlgait} extracting features from silhouette frames by a Restricted Boltzmann Machine. }

\pending{Regarding gait-based human identification in verification mode, Wu~\etal~\cite{wu2016comprehensive} proposed a 3D CNN network with triplets of continuous gait silhouettes as the input. They also proposed LBNet to learn paired filters on a pair of GEIs at the initial input layer and to output a similarity of two input GEIs. Furthermore, a comprehensive discussion was proposed by Takemura~\cite{2017inputoutput}, deeply investigating deep gait recognition in verification versus identification mode with contrastive loss and triplet ranking loss, respectively. The study found that triplet ranking loss could effectively improve cross-view accuracy from consideration of the intra-subject spatial displacement caused by view-invariant. Although the aforementioned divided early fusion models significantly improved performance, these methods~\cite{shiraga2016geinet,wu2016comprehensive,firstdlgait} based on GEI still suffer from performance degradation when evaluated on the clothes-changing subsets. Therefore, Yao~\cite{SGEI} introduced a novel gait template called skeleton gait energy image. This method could perform better in an unconstrained environment with changing views and clothes by averaging sequential skeletons instead of silhouettes to generate the GEI. Though divided early fusion models have significantly boosted recognition accuracy by applying spatiotemporal feature fusion at the input level, more subsequent work~\cite{bblin3d,2019_joint_zhangyuqi} indicated that the methods with spatiotemporal fusion at feature level capture more informative gait features spatial-temporally.}

\vspace{3mm} \noindent{\textbf{{Divided Late Fusion (S+T) Model}}}

\pending{\vspace{1mm} As named, divided late fusion models learn gait appearance before modeling and aggregating motion. This architecture advanced in fine-grained feature extraction on frame-level gait input rather than gait template, which helps to gain performance improvement and preserve spatial gait patterns in detail. Besides, the divided late fusion architecture fusing spatial-temporal features separately in two stages can help the community to investigate and understand human gait spatiotemporally better. With these advantages, many works have emerged for deep gait recognition.}

\pending{Regarding spatial information modeling in the context of divided late fusion architecture, 2D convolution neural networks are typically the first choice to encode features from structured inputs, including RGB images, silhouettes, depths, and infrared images. A milestone of the 2D CNN-based method, \ie GaitSet~\cite{gaitsetv1}, established a light and efficient network consisting of six convolutions along LeakyReLU activations and two max-pooling layers, which first achieved 70.3\% cross-view recognition accuracy on CASIA-B dataset under clothes changing condition. The following works utilized such GaitSet-like backbone integrated with local~\cite{gaitpart}, global-local~\cite{Lin_2021_ICCV}, and multi-scale representation learning strategies~\cite{gpgait}, greatly boosting gait recognition performance on the laboratory datasets. However, these models are light and shadow, suffering a significant degradation when the models are employed from indoor to outdoor scenes. Fan~\etal~\cite{fan2023opengait,fan2023deepgaitv2} contributed to such performance degradation because shadow architectures are insufficient to learn discriminative features from unprecedented challenges within real-world scenes, such as complex backgrounds, harmful occlusion, unpredictable illumination, arbitrary viewpoints, and diverse clothing changes. Therefore, GaitBase~\cite{fan2023opengait} with residual learning deepened networks and improved performance on in-the-wild datasets significantly. DeepGaitV2-2D~\cite{fan2023deepgaitv2} further increased the depth of networks and demonstrated the recognition accuracy was positively relevant to the depth of models on in-the-wild datasets. 
In addition to convolutional neural networks as spatial feature encoders for structured inputs like silhouettes and RGB images, other networks such as graph networks~\cite{teepe2021gaitgraph} and multilayer perceptron networks~\cite{shen2023lidargait} are used to encode gait features for different inputs, including skeletons and point clouds, respectively.}

\pending{For temporal information modeling in the context of divided late fusion architecture, there are two main categories of temporal modeling methods: parameter-free and parameter-based. On the one hand, parameter-free methods use statistical functions to aggregate temporal information among sequences into sequence-level features. Temporal average pooling~\cite{gaitsetv2} is a common example of this approach, which calculates the average value of each feature over the entire temporal dimension of the sequence. Temporal max pooling~\cite{gaitsetv1} is another statistical function that summarizes temporal information with superior recognition performance. However, they may not capture complex temporal dynamics and variations in the gait patterns as effectively as parameter-based methods.
On the other hand, parameter-based methods use learnable parameters to capture temporal dependencies within the sequence data. These methods often incorporate recurrent neural networks, 1D convolutions~\cite{gaitpart}, and attention mechanisms~\cite{dou2022metagait} to model the temporal dynamics of the gait sequence. While some works utilize learnable parameters to capture either long-term or short-term temporal features, recent work~\cite{Huang_2021_ICCV} suggests that multi-scale temporal feature learning with long-term, short-term, and frame-level features can better distinguish gait variants at different scales of temporal clues. Though these parameter-based methods help models learn informative gait dynamics, they come at the cost of increased computational complexity, which may not be practical for many real-world applications and even result in inferior performance compared to parameter-free methods~\cite{zheng2022gait3d,shen2023lidargait,li2023CCPG}.}

\vspace{3mm} \noindent{\textbf{{Joint Fusion (ST) Model}}}

\pending{\vspace{1mm} Compared to divided fusion models that learn space and time dimensions independently, joint fusion models consider both dimensions simultaneously in the learning process. Therefore these models can potentially capture more complex spatiotemporal patterns in gait sequences, leading to improved performance.}

\pending{To learn spatiotemporal features from structural gait sequences, such as depths and images, 3D CNNs~\cite{3dcnnAction} are ideal processing units, as they understand sequential sequences as a 3D tensor with the shape of $C \times T \times H \times W$, where $T$ represents the number of frames, $H$ and $W$ are the spatial dimensions, and $C$ refers to the number of channels.  The seminal work using the 3D convolutional network (C3D) for gait recognition is~\cite{first3dcnngait}. While inspiring, C3D is computationally intensive and hard to optimize on small-scale datasets. To address this, Lin~\etal~\cite{bblin3d,Lin_2021_ICCV} factorized a 3D kernel (\eg $3 \times 3 \times 3$) into two separate operations, a 2D convolution (\eg $1 \times 3 \times 3$) for spatial pattern modeling and a 1D convolution (\eg $3 \times 1 \times 1$) to construct temporal connections. Following this line, DeepGaitV2-P3D~\cite{fan2023deepgaitv2} compared 3D residual convolutions (R3D) and Pseudo 3D residual convolutions (P3D)~\cite{qiu2017p3d} and found that two types of 3D convolutions achieve comparative performance, but P3D has fewer learnable parameters (11.1 vs. 27.5M) and lower computational cost (2.9 vs. 6.8GFLOPs). Additionally, DeepGaitV2-P3D greatly improved the recognition accuracy of DeepGaitV2-2D by up to 9.1\% on various datasets with only a slight increase in parameters and computational cost.}

\pending{In addition to 3D convolution, there are many alternatives for performing joint spatial-temporal modeling. Recently, temporal shift operation~\cite{zheng2022gaitmultihop} further enhances temporal information modeling among neighboring frames. These methods are computationally efficient and do not require additional parameter tuning, making them popular in many gait recognition applications. Similarly, DyGait~\cite{wang2023dygait} and SiMo~\cite{2023ICASSPfudan_motionmatter} explicitly model human dynamical motion with prior knowledge by facilitating the temporal difference operation between frames. Recently, many works have explored transformers~\cite{2023ICASSPfudan_cotr} and spatial-temporal graph convolutional networks to model gait features from nonstructural inputs such as skeletons and point clouds.}

\subsubsection{Multi-modal architecture}

\pending{To construct discriminative deep features for gait recognition, alternative architectures explore the idea of utilizing multiple modalities. Unlike the aforementioned single-modality models that rely on architecture design to learn different semantic information, multi-modal methods explicitly utilize diverse semantic information from many input streams. In some literature, multi-modal architecture is defined as networks with multiple branches. In this paper, the term multi-modal models refers specifically to networks that utilize multiple modalities. Therefore a single modality having multiple pathways like GaitSet~\cite{gaitsetv1} is categorized into the single-modal model. }

\pending{To capture motion features of human gait, considering optical flow appears to be a reasonable solution. The first attempt is introduced by the seminal paper on Two-Stream Networks (TSN)~\cite{simonyan2014TSN} for human action recognition, where TSN adds a second pathway to learn the temporal information in a video by training a CNN on the optical flow stream. Following this trend, Ruben~\etal~\cite{delgado2021multimodal} extends to add a third path to capture structural information on the depth images. Furthermore, UGaitNet~\cite{marin2021ugaitnet} inherits the multimodal recognition framework and further discusses a neglected problem within multimodal gait recognition that some modalities can be missing at test time. To tackle such a problem, UGaitNet integrated a gate mechanism and a merge operation to combine information from modality-specialized branches. In that way, when a modality is missing, the gate can disable the input of such modality to its corresponding branch to not penalize the performance. }

\pending{To fully use body shape and structure information from the given modalities, the multi-modal methods integrating silhouettes and skeletons~\cite{2023ICASSPfudan_cotr,twostream_skeleton_pyj,twostream_skeleton_wlk} are the most commonly used. Precisely, silhouettes maintain body shape information but are sensitive to the factors of changing human appearance, such as clothing and camera viewpoints. Skeletons are robust to the viewpoints but ignore the body shape details. GaitHybrid~\cite{cai2021hybrid} was one of the first multi-stream networks that proposed a hybrid silhouette-skeleton body representation for gait recognition. Differently, GaitHybrid employed CNN on the skeleton heatmap to capture body structure information instead of unitizing GCN on the raw skeletons sequence with a shape of $C \times T \times N$, where N denotes the number of body key points, $T$ represents the number of frames, and $C$ refers to the number of channels of body keypoints. Typically, the silhouette-skeleton models as proposed in~\cite{twostream_skeleton_pyj,twostream_skeleton_wlk} consist of a pose feature encoder and a silhouette feature encoder independently for two modalities. Instead of obtaining human structural information from human skeletons, recent methods target learning 3D body shapes from diverse modalities, including 3D human mesh~\cite{hmrgait}, depth images~\cite{marin2021ugaitnet}, and 3D point clouds~\cite{han2022licamgait}.}

\subsection{Generative Model}
Unlike discriminative models that learn decision boundaries to make predictions, deep generative models learn the actual data distribution via reproducing samples as closely as possible from the given inputs~\cite{yu2017gaitgan,densegan}. Specifically, generative models summarize input data distribution and synthesize samples in the given data distribution. In the context of gait recognition, deep generative models can be used to eliminate the challenges of gait appearance changing caused by various real-world factors, including object carrying~\cite{2020cvpr2}, clothes changing~\cite{zhang2020gaitnet}, and diverse viewpoints observing~\cite{densegan}. Generally, generative models for gait recognition can be divided into two main categories. One is the deep autoencoder network, constructing probability distribution of given data based on maximum likelihood estimation. The other one is the generative adversarial network, finding the Nash equilibrium of the minimax game between the data generator and sample discriminator. In the following parts, we provide a detailed introduction to these two models for generative gait models.

\subsubsection{Deep autoencoders}
\pending{Deep autoencoders (DAEs) are a specific type of neural network architecture that aims to learn compact representations, also known as latent representations. The main objective of DAEs is to encode the input data into a lower-dimensional feature representation using an encoder. Then DAEs decode it back to reconstruct the original input using a decoder. The encoder typically consists of fully connected and/or convolutional layers, while the decoder performs the inverse operations. By minimizing the reconstruction error, which quantifies the dissimilarity between the input and reconstructed data, DAEs effectively learn latent features. These latent features serve as concise representations that capture the essential characteristics of the original data and can be extracted for various applications, including tasks such as generating new samples or performing gait recognition. Following this line, Yu~\etal~\cite{2017shiqiyuvae} proposed a uniform model named SPAE that adopted autoencoders to learn invariant gait features for gait recognition. With the stacked autoencoders, SPAE encodes the given GEIs from various viewpoints with carrying and clothing conditions from the input layers, and then constructs GEIs from the side view with the normal condition. With a similar spirit, Babaee~\etal~\cite{babaee2019vae} proposed an incomplete to complete GEI converter, which is able to transform incomplete GEI or even a frame of silhouette to the corresponding complete GEI. Furthermore, recent methods integrated disentangled representation learning with encoder-decoder architecture. For instance, Li~\etal~\cite{2020cvpr2} proposed ICDNet that consisted of an encoder to disentangle latent identity and covariant features from an input GEI and a decoder to perform two reconstructions. Zhang~\etal~\cite{zhang2019gait} disentangled gait features in two components: pose and pose-irrelevant, while the improved version~\cite{zhang2020gaitnet} further decoupled the pose-irrelevant features into discriminative canonical feature and appearance feature, improving gait recognition accuracy and providing interoperability. However, DAEs suffer from posterior collapse and imperfect reconstruction, resulting in uninformative latent space learning and blur sample synthesis, respectively. }

\subsubsection{Generative adversarial network}
\pending{Generative Adversarial Networks (GANs) are representative generative models with a strong performance in data synthesis. These generative methods can effectively generate data with high quality and perverse identity consistency at the same time, which makes them very popular in the context of gait recognition for data generation. Generally, GANs involve two distinct networks, where a \textit{generator} receives a latent code sampled from a prior distribution as input to generate data, and a \textit{discriminator} distinguishes between real data and the synthesized data. Going through the zero-sum game between generator and discriminator, GANs can enjoy high-fidelity generation. Thus, GANs are gradually employed to solve many challenges of gait recognition, \ie differences in viewpoints, diverseness of clothing, and domain gap between different datasets. To perform gait recognition across two arbitrary views, GANs transform gaits into either a unified view or an identical view to the reference samples. For instance, Yu~\etal~\cite{yu2017gaitgan} proposed a generative adversarial network called GaitGAN, which can transform the gait data from any view, clothing, and carrying conditions to the canonical side view of a walking gait in normal clothing without carrying objects. Nevertheless, it is a challenge to preserve human identity information when performing appearance transformation via generative models. To this end, GaitGAN adopts the identification discriminator to constrain the generated gait samples preserving identical identity information to the source gait samples, while GaitGANv2 further integrates a multi-loss strategy to enhance the preservation of human identity features further. A recent PSTN~\cite{2020PSTN} had integrated a prior pairwise spatial transformer ahead of the recognition network, recusing spatial feature misalignment to enable recognition more robust on cross-view scenarios. Instead of focusing on the viewpoint transformation, Frame-GAN proposes to deal with the realistic challenge of low frame rate on gait recognition and then helps to improve gait recognition performance by increasing more gait frames. Toward gait recognition at low frame rate inputs, Xu~\etal proposed a method named PA-GCRNet~\cite{xuchigaitgan} that can even reconstruct a complete gait cycle of silhouettes. }

\pending{The generative models have demonstrated the necessity to eliminate the challenges of various real-world factors. However, most generative methods heavily relied on supervised labels for training, lacking flexibility for practical usage. Recently GaitEditor~\cite{gaiteditor} has adopted the GAN inversion technique to construct latent space with semantic separability, enabling multiple gait attribute manipulation. It shows the great potential of generative gait models with unsupervised learning toward generating a broad range of realistic attributes such as viewpoints, clothing, walking speed, gender, and even age.}

\section{Datasets and Evaluation}

In this section, we first introduce the public gait datasets we can find in the literature to help researchers find their research of interest. Then we evaluate recent advances in the representative datasets to illustrate the development of deep gait models.

\subsection{Datasets for Gait Recognition}

 We list all public gait datasets that we can find in the literature in Table~\ref{tab:dataset}. It is concluded that large-scale gait datasets typically have limited attributes, while those with considerable attributes are small-scale. Large-scale gait datasets are essential to provide statistical evaluation for gait recognition. Especially in these years, the increasing data size and deep learning greatly advance performance. However, collecting a large-scale gait dataset requires much more time, storage, and cost than a similar-sized face or fingerprint dataset.
 
 To this end, one possible solution is that the research community can work together to collect data and train methods using federated learning or other privacy-protecting learning methods. The alternative is to create synthetic gait datasets using virtual human body models, which is also an exciting and promising solution. With concerns about data privacy and security, collecting a large gait dataset is much more challenging nowadays. Some laws, such as \textit{European General Data Protection Regulation} (GDPR)~\cite{gdpr} and the \textit{Data Security Law of the P. R. of China}~\cite{law-dsl}, put substantial restrictions to protect our privacy and improve data security in data collection and usage. It is a challenge and a new opportunity for the academic community to develop better methods to protect our privacy and improve the security of our society. 
 
\subsection{Performance Comparisons}
Comparing deep gait models on diverse datasets is not easy since there are over 60 datasets, as reported in Table~\ref{tab:dataset}, and each dataset might have diverse criteria and evaluation protocols. To present clearly, we select some representative datasets and report the state-of-the-art methods on these popular datasets, aiming to draw valuable observations from the performance comparison and analysis. \pending{To provide a comprehensive performance comparison on deep gait recognition, we evaluate deep gait recognition methods from three scenarios: (1) \textit{cross-view scenarios}, focusing on evaluation on two popular datasets, CASIA-B~\cite{casiab} and OUMVLP~\cite{dataset2017oumvlp}. (2) \textit{in-the-wild scenarios}, reporting results on two popular outdoor datasets, Gait3D~\cite{zheng2022gait3d} and GREW~\cite{dataset2021grew}. (3) \textit{Clothes-changing scenarios}, relating to three outdoor datasets with clothing change attributes, CASIA-E~\cite{song2022casiae}, CCPG~\cite{li2023CCPG}, and FVG~\cite{zhang2020gaitnet}. (4) \textit{Point cloud-based scenarios}, focusing on gait recognition using 3D point clouds and evaluating on two LiDAR-based datasets, SUSTech1K~\cite{shen2023lidargait} and LiCamGait~\cite{han2022licamgait}.}

\begin{table}[htbp]
\label{tab:indoordatasets}
\caption{\pending{Cross-view recognition performance on two popular datasets, CASIA-B~\cite{casiab} and OUMVLP~\cite{dataset2017oumvlp}.}}
\scalebox{0.72}{
\begin{tabular}{l|l|l|llll|c}
\hline
\multirow{2}{*}{Input}        & \multirow{2}{*}{Method}   & \multirow{2}{*}{Publication} & \multicolumn{4}{c|}{\textbf{CASIAB}}                                                              & \textbf{OUMVLP} \\ \cline{4-8} 
                              &                           &                              & \multicolumn{1}{l|}{NM}   & \multicolumn{1}{l|}{BG}   & \multicolumn{1}{l|}{CL}   & Mean &  Mean      \\ \hline
\multirow{43}{*}{Silhouette} & GaitGAN~\cite{yu2017gaitgan}                 & CVPRW17                      & \multicolumn{1}{l|}{61.0}   & \multicolumn{1}{l|}{39.0}   & \multicolumn{1}{l|}{22.3} & 40.8 & -      \\ \cline{2-8} 
                              & ACL~\cite{2019zhangACL}                       & TIP19                        & \multicolumn{1}{l|}{96.0}   & \multicolumn{1}{l|}{-}    & \multicolumn{1}{l|}{-}    & -    & 89.0     \\ \cline{2-8} 
                              & Song~\cite{song2019gaitnet}              & PR19                         & \multicolumn{1}{l|}{89.9} & \multicolumn{1}{l|}{-}    & \multicolumn{1}{l|}{-}    & -    & -      \\ \cline{2-8} 
                              & Zhang~\etal~\cite{2019_joint_zhangyuqi}                & PR19                         & \multicolumn{1}{l|}{91.2} & \multicolumn{1}{l|}{75.0}   & \multicolumn{1}{l|}{54.0}   & 73.4 & -       \\ \cline{2-8} 
                              & GaitSet~\cite{gaitsetv1}                   & AAAI19                       & \multicolumn{1}{l|}{95.0}   & \multicolumn{1}{l|}{87.2} & \multicolumn{1}{l|}{70.4} & 84.2 & 87.1   \\ \cline{2-8} 
                              & EV-Gait~\cite{evgait}                   & CVPR19                       & \multicolumn{1}{l|}{94.1} & \multicolumn{1}{l|}{-}    & \multicolumn{1}{l|}{-}    & -    & -      \\ \cline{2-8} 
                              & GaitPart~\cite{gaitpart}                  & CVPR20                       & \multicolumn{1}{l|}{96.2} & \multicolumn{1}{l|}{91.5} & \multicolumn{1}{l|}{78.7} & 88.8 & 88.7   \\ \cline{2-8} 
                              & GLN~\cite{gln}                       & ECCV20                       & \multicolumn{1}{l|}{96.9} & \multicolumn{1}{l|}{94.0}   & \multicolumn{1}{l|}{77.5} & 89.5 & 89.2   \\ \cline{2-8} 
                              & MT3D~\cite{bblin3d}                      & MM20                         & \multicolumn{1}{l|}{96.7} & \multicolumn{1}{l|}{93.0}   & \multicolumn{1}{l|}{81.5} & 90.4 & -       \\ \cline{2-8} 
                              & Dresser~\cite{wu2020condition}                   & TIP20                        & \multicolumn{1}{l|}{94.8} & \multicolumn{1}{l|}{88.8} & \multicolumn{1}{l|}{81.8} & 88.5 & -       \\ \cline{2-8} 
                              & PSTN~\cite{2020PSTN}                      & TCSVT20                      & \multicolumn{1}{l|}{92.7} & \multicolumn{1}{l|}{-}    & \multicolumn{1}{l|}{-}    & -    & 63.1   \\ \cline{2-8} 
                              & PA-GCRNet~\cite{xuchigaitgan}                 & ECCV20                       & \multicolumn{1}{l|}{-}    & \multicolumn{1}{l|}{-}    & \multicolumn{1}{l|}{-}    & 74.7 & -      \\ \cline{2-8} 
                              & DV-GEIs~\cite{liao2020dvgei}                   & IJCB20                       & \multicolumn{1}{l|}{83.4} & \multicolumn{1}{l|}{-}    & \multicolumn{1}{l|}{-}    & -    & -      \\ \cline{2-8} 
                              & Yao~\etal~\cite{yao2022TMM}                  & TMM2022                      & \multicolumn{1}{l|}{97.9} & \multicolumn{1}{l|}{95.5} & \multicolumn{1}{l|}{46.2} & 79.9 &        \\ \cline{2-8} 
                              & CSTL~\cite{Huang_2021_ICCV}                      & ICCV21                       & \multicolumn{1}{l|}{97.8} & \multicolumn{1}{l|}{93.6} & \multicolumn{1}{l|}{84.2} & 91.9 & 90.2   \\ \cline{2-8} 
                              & 3DLocal~\cite{3dlocal_2021_ICCV}                   & ICCV21                       & \multicolumn{1}{l|}{97.5} & \multicolumn{1}{l|}{94.3} & \multicolumn{1}{l|}{83.7} & 91.8 & 90.9   \\ \cline{2-8} 
                              & GaitGL~\cite{Lin_2021_ICCV}                    & ICCV21                       & \multicolumn{1}{l|}{97.4} & \multicolumn{1}{l|}{94.5} & \multicolumn{1}{l|}{83.6} & 91.8 & 89.7   \\ \cline{2-8} 
                              & SDML                      & ICASSP21                     & \multicolumn{1}{l|}{95.1} & \multicolumn{1}{l|}{89.2} & \multicolumn{1}{l|}{74.6} & -    & -      \\ \cline{2-8} 
                              & SelfGait~\cite{selfgait2021}                  & ICASSP21                     & \multicolumn{1}{l|}{93.5} & \multicolumn{1}{l|}{90.1} & \multicolumn{1}{l|}{81.3} & 88.3 & 89.9   \\ \cline{2-8} 
                              & GaitMask~\cite{lin2021gaitmask}                  & BMCV21                       & \multicolumn{1}{l|}{96.8} & \multicolumn{1}{l|}{93.2} & \multicolumn{1}{l|}{84.2} & 91.4 & 90.2   \\ \cline{2-8} 
                              & Koopman~\cite{zhang2021_CVPR_koopman}                   & CVPR21                       & \multicolumn{1}{l|}{-}    & \multicolumn{1}{l|}{-}    & \multicolumn{1}{l|}{-}    & -    & 74.7   \\ \cline{2-8} 
                              & Vi-GaitGL~\cite{chai2021view}                 & ICIP21                       & \multicolumn{1}{l|}{96.2} & \multicolumn{1}{l|}{92.9} & \multicolumn{1}{l|}{87.2} & 92.1 & 89.9   \\ \cline{2-8} 
                              & CapsNet~\cite{sepas2021gait_capsnet}                   & ICPR21                       & \multicolumn{1}{l|}{95.7} & \multicolumn{1}{l|}{90.7} & \multicolumn{1}{l|}{72.4} & 86.3 & -      \\ \cline{2-8} %
                              & SRN~\cite{hou2021gln_tbiom}                       & TBIOM21                      & \multicolumn{1}{l|}{97.1} & \multicolumn{1}{l|}{94.0}   & \multicolumn{1}{l|}{81.8} & 91.0   & 89.9   \\ \cline{2-8} 
                              & MvGGAN~\cite{chen2021mvggan}                    & TIP21                        & \multicolumn{1}{l|}{97.1} & \multicolumn{1}{l|}{91.9} & \multicolumn{1}{l|}{75.6} & 88.2 & 58.4   \\ \cline{2-8} 
                              & GaitMPL~\cite{dou2022gaitmpl}                   & TIP22                        & \multicolumn{1}{l|}{97.5} & \multicolumn{1}{l|}{94.5} & \multicolumn{1}{l|}{88.0}   & 93.3 & 89.6   \\ \cline{2-8} 
                              & Lagrange~\cite{Chai_2022_CVPR}                  & CVPR22                       & \multicolumn{1}{l|}{96.9} & \multicolumn{1}{l|}{93.5} & \multicolumn{1}{l|}{86.5} & 92.3 & 90.0     \\ \cline{2-8} 
                              & GaitSlice~\cite{li2022gaitslice}                 & PR22                         & \multicolumn{1}{l|}{96.7} & \multicolumn{1}{l|}{92.4} & \multicolumn{1}{l|}{81.6} & 90.2 & 89.3   \\ \cline{2-8} 
                              & MetaGait~\cite{dou2022metagait}                  & ECCV22                       & \multicolumn{1}{l|}{98.7} & \multicolumn{1}{l|}{96.0}   & \multicolumn{1}{l|}{89.3} & 94.7 & 92.4   \\ \cline{2-8} 
                              & STAR~\cite{huang2022star}                      & TBIOM22                      & \multicolumn{1}{l|}{97.3} & \multicolumn{1}{l|}{93.9} & \multicolumn{1}{l|}{84.0}   & 91.7 & 89.7   \\ \cline{2-8} 
                              & GaitStrip~\cite{wang2022gaitstrip}                 & ACCV22                       & \multicolumn{1}{l|}{97.6} & \multicolumn{1}{l|}{95.2} & \multicolumn{1}{l|}{86.2} & -    & -      \\ \cline{2-8} 
                              & GPAN~\cite{chen2022gaitGPAN}                      & TBIOM22                      & \multicolumn{1}{l|}{97.7} & \multicolumn{1}{l|}{94.2} & \multicolumn{1}{l|}{81.8} & 91.2 & 81.7   \\ \cline{2-8} 
                              & ESNet~\cite{huang2022ESNet}                     & TCSVT22                      & \multicolumn{1}{l|}{97.4} & \multicolumn{1}{l|}{94.0}   & \multicolumn{1}{l|}{84.0}   & 91.8 & 89.4   \\ \cline{2-8} 
                              & GQAN~\cite{hou_2022_quality}                      & TNNLS22                      & \multicolumn{1}{l|}{98.5} & \multicolumn{1}{l|}{95.4} & \multicolumn{1}{l|}{84.5} & 92.8 & 96.16  \\ \cline{2-8} 
                              & GaitGCI~\cite{dou2023gaitgci}                   & CVPR23                       & \multicolumn{1}{l|}{98.4} & \multicolumn{1}{l|}{96.6} & \multicolumn{1}{l|}{88.5} & 94.5 & 92.1   \\ \cline{2-8} 
                              & DANet~\cite{ma2023dynamic_CVPR}                     & CVPR23                       & \multicolumn{1}{l|}{98.0}   & \multicolumn{1}{l|}{95.9} & \multicolumn{1}{l|}{89.9} & 94.6 & 90.7   \\ \cline{2-8} 
                              & GaitBase~\cite{fan2023opengait}                  & CVPR23                       & \multicolumn{1}{l|}{97.6} & \multicolumn{1}{l|}{94.0}   & \multicolumn{1}{l|}{77.4} & 89.7 & 90.8   \\ \cline{2-8} 
                              & GaitCoTr~\cite{2023ICASSPfudan_cotr}                  & ICASSP23                     & \multicolumn{1}{l|}{97.9} & \multicolumn{1}{l|}{95.0}   & \multicolumn{1}{l|}{89.4} & 94.1 & -      \\ \cline{2-8} 
                              & Li~\etal~\cite{2023ICASSPfudan_motionmatter}                   & ICASSP23                     & \multicolumn{1}{l|}{98.0}   & \multicolumn{1}{l|}{96.1} & \multicolumn{1}{l|}{88.1} & 94.1 & 90.5   \\ \cline{2-8} 
                              & Wang~\etal~\cite{twostream_skeleton_wlk}                 & ICASSP23                     & \multicolumn{1}{l|}{97.7} & \multicolumn{1}{l|}{93.8} & \multicolumn{1}{l|}{92.7} & 94.7 & 91     \\ \hline
\multirow{8}{*}{Skeleton}    & PoseGait~\cite{2020liao_posegait}                  & PR20                         & \multicolumn{1}{l|}{68.7} & \multicolumn{1}{l|}{44.5} & \multicolumn{1}{l|}{36.0}   & 49.7 & -      \\ \cline{2-8} 
                              & GaitGraph~\cite{teepe2021gaitgraph}                 & ICIP21                       & \multicolumn{1}{l|}{87.7} & \multicolumn{1}{l|}{74.8} & \multicolumn{1}{l|}{66.3} & 76.3 & 20.4   \\ \cline{2-8} 
                              & GaitGraph2~\cite{teepe2022gaitgraph2}                & CVPRW22                      & \multicolumn{1}{l|}{82.0}   & \multicolumn{1}{l|}{73.2} & \multicolumn{1}{l|}{63.6} & 72.9 & 67.1   \\ \cline{2-8} 
                              & CNN-Pose~\cite{2020anPoseDataset}                  & TBIOM20                      & \multicolumn{1}{l|}{-}    & \multicolumn{1}{l|}{-}    & \multicolumn{1}{l|}{-}    & -    & 20.4   \\ \cline{2-8} 
                              & GaitTAKE~\cite{hsu2022gaittake}                  & ICIP22                       & \multicolumn{1}{l|}{98.0}   & \multicolumn{1}{l|}{97.5} & \multicolumn{1}{l|}{92.2} & 95.9 & 90.4   \\ \cline{2-8} 
                              & GaitPoint~\cite{chen2022gaitpoint}                 & ICIP22                       & \multicolumn{1}{l|}{93.0}   & \multicolumn{1}{l|}{92.0}   & \multicolumn{1}{l|}{81.1} & 88.7 & -      \\ \cline{2-8} 
                              & GaitMixer~\cite{pinyoanuntapong2023gaitmixer}                 & ICASSP23                     & \multicolumn{1}{l|}{94.9} & \multicolumn{1}{l|}{85.6} & \multicolumn{1}{l|}{84.5} & 88.3 & -      \\ \cline{2-8} 
                              & MMGaitFormer~\cite{CVPR_23_MMGaitFormoer}              & CVPR23                       & \multicolumn{1}{l|}{98.4} & \multicolumn{1}{l|}{96.0}   & \multicolumn{1}{l|}{94.8} & 96.4 & 90.1   \\ \hline
\multirow{5}{*}{RGB}          & ModelGait~\cite{hmrgait}                 & ACCV20                       & \multicolumn{1}{l|}{97.9} & \multicolumn{1}{l|}{93.1} & \multicolumn{1}{l|}{77.6} & 89.5 & 95.8   \\ \cline{2-8} 
                              & GaitNet~\cite{zhang2020gaitnet}                   & CVPR20                       & \multicolumn{1}{l|}{92.3} & \multicolumn{1}{l|}{88.9} & \multicolumn{1}{l|}{62.3} & 81.2 & -      \\ \cline{2-8} 
                              & MvModelGait~\cite{Li_2021_ICCV_mvmodelgait}               & ICCVW21                      & \multicolumn{1}{l|}{98.1} & \multicolumn{1}{l|}{93.4} & \multicolumn{1}{l|}{80.7} & 90.7 & 89.7   \\ \cline{2-8} 
                              & GaitEdge~\cite{liang2022gaitedge}                  & ECCV22                       & \multicolumn{1}{l|}{97.9} & \multicolumn{1}{l|}{96.1} & \multicolumn{1}{l|}{86.4} & 93.5 & -      \\ \cline{2-8} 
                              & Zhu~\etal~\cite{Zhu_2023_WACV}                 & WACV23                       & \multicolumn{1}{l|}{97.5} & \multicolumn{1}{l|}{94.8} & \multicolumn{1}{l|}{84.1} & 92.1 & 89.8   \\ \hline
\end{tabular}
}
\end{table}
\subsubsection{Evaluation on cross-view gait recognition}
\pending{To explore cross-view gait recognition, the Institute of Automation, Chinese Academy of Sciences, released the \textit{CASIA-B}~\cite{casiab} gait dataset in 2005, which is the first gait dataset containing over 100 subjects with walking sequences captured in dense viewpoints and diverse walking conditions \eg, walking normally (NM) or wearing a bag (BG) or a coat (CL). Alternatively, the Institute of Scientific and Industrial Research (\textit{ISIR}), Osaka University (\textit{OU}), also provided a cross-view dataset \textit{OUMVLP}~\cite{dataset2017oumvlp}~\cite{dataset2017oumvlp}. It is the largest multi-view gait dataset captured in the laboratory environment. OUMVLP contains 10307 subjects, and each participant contributes 28 sequences (7 cameras × 2 forward and backward × 2). }

\pending{As listed in Table~\ref{tab:indoordatasets}, many invaluable investigations have been proposed to tackle gait recognition toward cross-view scenarios. We can make three observations from the performance evaluation: (1) The deep models have progressively upgraded cross-view accuracy in both CASIA-B and OUMVLP datasets. The state-of-the-art methods~\cite{fan2023opengait,wang2023dygait,ma2023dynamic_CVPR} achieve over 90\% rank-1 accuracy on two datasets under cross-view evaluation protocol. (2) The silhouette-based methods dominated the field of study, while methods taking other modalities ( \ie RGB images and skeletons) are relatively lacking in the study. (3) These models behave differently on two datasets. In other words, some methods make comparable performance on CASIA-B, while there is a clear performance gap when implemented in OUMVLP. For example, GaitGCI~\cite{dou2023gaitgci} and DANet~\cite{ma2023dynamic_CVPR} achieve 94.5\% and 94.6\%, respectively, on CASIA-B. For OUMVLP, GaitGCI~\cite{dou2023gaitgci} outperforms DANet by 1.4\% rank-1 recognition accuracy (92.1\% v.s. 90.7\%).}

\pending{With the aforementioned observations, we can conclude that (1) indoor cross-view gait recognition turns mature, and it suggests diverting gait recognition into more challenging settings like outdoor scenarios. (2) Recognizing pedestrians via other modalities is potential to investigate further. (3) For deep gait models, it is suggested to conduct statistically significant conclusions from the evaluation on a large-scale dataset.}

\begin{table}[htbp]
\centering
\label{tab:wilddatasets}
\caption{\pending{Gait recognition performance on two popular in-the-wild datasets, GREW and Gait3D.}}
\scalebox{0.65}{
\begin{tabular}{l|l|llll|llll}
\hline
\multirow{2}{*}{Method} & \multirow{2}{*}{Venue} & \multicolumn{4}{c|}{\textbf{Gait3D}~\cite{zheng2022gait3d}}                                                                  & \multicolumn{4}{c}{\textbf{GREW}~\cite{dataset2021grew}}                                                                      \\ \cline{3-10} 
                        &                              & \multicolumn{1}{l|}{R@1} & \multicolumn{1}{l|}{R@5} & \multicolumn{1}{l|}{mAP}   & mINP  & \multicolumn{1}{l|}{R@1} & \multicolumn{1}{l|}{R@5} & \multicolumn{1}{l|}{R@10} & R@20 \\ \hline
GEINet~\cite{shiraga2016geinet}                  & ICB16                        & \multicolumn{1}{l|}{5.4}   & \multicolumn{1}{l|}{14.2}  & \multicolumn{1}{l|}{5.1}  & 3.14  & \multicolumn{1}{l|}{6.8}  & \multicolumn{1}{l|}{13.4} & \multicolumn{1}{l|}{17.0}  & 21.0  \\ \hline
GaitSet~\cite{gaitsetv1}                 & AAAI19                       & \multicolumn{1}{l|}{36.7}  & \multicolumn{1}{l|}{58.3}  & \multicolumn{1}{l|}{30}    & 17.3  & \multicolumn{1}{l|}{46.3}  & \multicolumn{1}{l|}{63.6}  & \multicolumn{1}{l|}{70.3}   & 76.8   \\ \hline
GaitPart~\cite{gaitpart}                & CVPR20                       & \multicolumn{1}{l|}{28.2}  & \multicolumn{1}{l|}{47.6}  & \multicolumn{1}{l|}{21.6}  & 12.4  & \multicolumn{1}{l|}{44}    & \multicolumn{1}{l|}{60.7}  & \multicolumn{1}{l|}{67.3}   & 73.5   \\ \hline
GLN~\cite{gln}                     & ECCV20                       & \multicolumn{1}{l|}{31.4}  & \multicolumn{1}{l|}{52.9}  & \multicolumn{1}{l|}{24.7}  & 13.6  & \multicolumn{1}{l|}{-}     & \multicolumn{1}{l|}{-}     & \multicolumn{1}{l|}{-}      & -      \\ \hline
PoseGait~\cite{2020liao_posegait}                & PR20                         & \multicolumn{1}{l|}{0.2}   & \multicolumn{1}{l|}{1.1}   & \multicolumn{1}{l|}{0.5}   & 0.3   & \multicolumn{1}{l|}{0.2}   & \multicolumn{1}{l|}{1.1}   & \multicolumn{1}{l|}{2.2}    & 4.3    \\ \hline
GaitGraph~\cite{teepe2021gaitgraph}               & ICIP21                       & \multicolumn{1}{l|}{6.3}   & \multicolumn{1}{l|}{16.2}  & \multicolumn{1}{l|}{5.2}   & 2.4   & \multicolumn{1}{l|}{1.3}   & \multicolumn{1}{l|}{3.5}   & \multicolumn{1}{l|}{5.1}    & 7.5    \\ \hline
CSTL~\cite{Huang_2021_ICCV}                    & ICCV21                       & \multicolumn{1}{l|}{11.7}  & \multicolumn{1}{l|}{19.2}  & \multicolumn{1}{l|}{5.6}   & 2.6   & \multicolumn{1}{l|}{-}     & \multicolumn{1}{l|}{-}     & \multicolumn{1}{l|}{-}      & -      \\ \hline
GaitGL~\cite{Lin_2021_ICCV}                  & ICCV21                       & \multicolumn{1}{l|}{63.8}  & \multicolumn{1}{l|}{80.5}  & \multicolumn{1}{l|}{55.9} & 36.7 & \multicolumn{1}{l|}{68}    & \multicolumn{1}{l|}{80.7}  & \multicolumn{1}{l|}{85}     & 88.2   \\ \hline
SMPLGait~\cite{zheng2022gait3d}                & CVPR22                       & \multicolumn{1}{l|}{46.3}  & \multicolumn{1}{l|}{64.5}  & \multicolumn{1}{l|}{37.2}  & 22.2  & \multicolumn{1}{l|}{-}     & \multicolumn{1}{l|}{-}     & \multicolumn{1}{l|}{-}      & -      \\ \hline
GaitGCI~\cite{dou2023gaitgci}                 & CVPR23                       & \multicolumn{1}{l|}{50.3}  & \multicolumn{1}{l|}{68.5}  & \multicolumn{1}{l|}{39.5}  & 24.3  & \multicolumn{1}{l|}{68.5}  & \multicolumn{1}{l|}{80.8}  & \multicolumn{1}{l|}{84.9}   & 87.7   \\ \hline
GaitBase~\cite{fan2023opengait}                & CVPR23                       & \multicolumn{1}{l|}{64.6}  & \multicolumn{1}{l|}{-}     & \multicolumn{1}{l|}{-}     & -     & \multicolumn{1}{l|}{60.1}  & \multicolumn{1}{l|}{-}     & \multicolumn{1}{l|}{-}      & -      \\ \hline
DANet~\cite{ma2023dynamic_CVPR}                   & CVPR23                       & \multicolumn{1}{l|}{48}    & \multicolumn{1}{l|}{69.7}  & \multicolumn{1}{l|}{-}     & -     & \multicolumn{1}{l|}{-}     & \multicolumn{1}{l|}{-}     & \multicolumn{1}{l|}{-}      & -      \\ \hline
RealGait~\cite{zhang2022realgait}                & Arxiv23                      & \multicolumn{1}{l|}{-}     & \multicolumn{1}{l|}{-}     & \multicolumn{1}{l|}{-}     & -     & \multicolumn{1}{l|}{54.1} & \multicolumn{1}{l|}{71.5} & \multicolumn{1}{l|}{77.6}  & 81.7  \\ \hline
DyGait~\cite{wang2023dygait}                  & Arxiv23                      & \multicolumn{1}{l|}{66.3}  & \multicolumn{1}{l|}{80.8}  & \multicolumn{1}{l|}{56.4}  & 37.3  & \multicolumn{1}{l|}{71.4}  & \multicolumn{1}{l|}{83.2}  & \multicolumn{1}{l|}{86.8}   & 89.5   \\ \hline
MTSGait~\cite{zheng2022gaitmultihop}                 & MM22                         & \multicolumn{1}{l|}{48.7}  & \multicolumn{1}{l|}{67.1}  & \multicolumn{1}{l|}{37.6} & 21.9 & \multicolumn{1}{l|}{55.32} & \multicolumn{1}{l|}{71.3} & \multicolumn{1}{l|}{76.9}  & 81.6  \\ \hline
DeepGaitV2~\cite{fan2023deepgaitv2}              & Arxiv23                      & \multicolumn{1}{l|}{74.4}  & \multicolumn{1}{l|}{88}    & \multicolumn{1}{l|}{65.8}  &       & \multicolumn{1}{l|}{77.7}  & \multicolumn{1}{l|}{87.9}  & \multicolumn{1}{l|}{90.6}   &        \\ \hline
GaitRef~\cite{zhu2023gaitref}                 & Arxiv23                      & \multicolumn{1}{l|}{49}    & \multicolumn{1}{l|}{69.3}  & \multicolumn{1}{l|}{40.7} & 25.3 & \multicolumn{1}{l|}{53}    & \multicolumn{1}{l|}{67.9}  & \multicolumn{1}{l|}{73}     & 77.5   \\ \hline
GaitCoTr~\cite{2023ICASSPfudan_cotr}                & ICASSP23                     & \multicolumn{1}{l|}{-}     & \multicolumn{1}{l|}{-}     & \multicolumn{1}{l|}{-}     & -     & \multicolumn{1}{l|}{55.6}  & \multicolumn{1}{l|}{70.9}  & \multicolumn{1}{l|}{76.2}   & 80.4   \\ \hline
GPGait~\cite{gpgait}                  & Arxiv23                      & \multicolumn{1}{l|}{22.5}  & \multicolumn{1}{l|}{}      & \multicolumn{1}{l|}{}      & -     & \multicolumn{1}{l|}{53.6} & \multicolumn{1}{l|}{-}     & \multicolumn{1}{l|}{-}      & -      \\ \hline
\end{tabular}
}
\end{table}

\subsubsection{Evaluation on in-the-wild gait recognition}
\pending{Although great achievements have been made in indoor laboratory datasets, gait recognition would encounter more challenges in real-world environments. To delve into this open problem, Gait3D and GREW were constructed to promote gait recognition in real-world scenes. 
For Gait3D, it collected gait data from 39 cameras in a supermarket, which constructed a cross-camera gait recognition dataset with 4000 identities and 25,309 sequences in total. GREW is the in-the-wild dataset with the most identities by far. It contains 26,345 subjects and 128,671 sequences from 882 cameras. Besides, it also provides 233,857 sequential distractors.}

\pending{From the analysis presented in Table~\ref{tab:wilddatasets}, several important observations can be made. Firstly, there is a significant drop in performance when deep gait methods are applied to outdoor datasets compared to indoor datasets~\cite{fan2023opengait}. Secondly, silhouette-based methods outperform those using estimated skeletons by a considerable margin~\cite{gpgait}. Thirdly, deep models show better generalization performance on the GREW dataset, despite it having a larger number of identities compared to Gait3D~\cite{fan2023deepgaitv2}. 
These observations reflect that the challenges present in unconstrained environments, such as occlusion~\cite{dataset_tum_iitkgp}, diverse viewpoints~\cite{song2022casiae}, and poor resolution~\cite{dataset2021grew}, have a detrimental effect on the performance of gait recognition models. These factors highlight the impact of realistic factors on the generation of robust gait representations. It is challenging to obtain ideal silhouettes and skeletons as observed in laboratory settings. For the results that larger scale of the dataset but have better performance, we hypothesize that the limited diversity of viewpoints in GREW, including front-view, side-view, and back-view, contributes to the improved performance of deep models. On the other hand, Gait3D suffers from spatial misalignment due to the collection of gait data from various vertical viewpoints and angles. The analysis of the raw silhouettes from both datasets supports the hypothesis that the diversity of viewpoints in the GREW dataset contributes to its better performance than Gait3D.}

\begin{table}[htbp]
\label{tab:clothesdataset}
\centering
\caption{\pending{Gait recognition performance on three clothes-changing datasets, FVG~\cite{zhang2020gaitnet}, CASIA-E~\cite{song2022casiae}, and CCPG~\cite{li2023CCPG}.}}
\scalebox{0.75}{
\begin{tabular}{lccccccc}
\hline
\multicolumn{8}{c}{\textbf{FVG~\cite{zhang2020gaitnet}}}                                                                                                                                                                                                \\ \hline
\multicolumn{2}{c|}{}                                             & \multicolumn{3}{c|}{CL}                                                            & \multicolumn{3}{c}{ALL}                                       \\ \hline
\multicolumn{1}{l|}{GEINet~\cite{shiraga2016geinet}}     & \multicolumn{1}{c|}{ICB16}      & \multicolumn{3}{c|}{6.5}                                                           & \multicolumn{3}{c}{13}                                        \\ \hline
\multicolumn{1}{l|}{LBNet~\cite{wu2016comprehensive}}      & \multicolumn{1}{c|}{T-PAMI16}   & \multicolumn{3}{c|}{23.2}                                                          & \multicolumn{3}{c}{40.7}                                      \\ \hline
\multicolumn{1}{l|}{GaitNetv1~\cite{gaitsetv1}}  & \multicolumn{1}{c|}{CVPR19}     & \multicolumn{3}{c|}{56.8}                                                          & \multicolumn{3}{c}{81.2}                                      \\ \hline
\multicolumn{1}{l|}{GaitNetv2~\cite{gaitsetv2}}  & \multicolumn{1}{c|}{GaitNetv2}  & \multicolumn{3}{c|}{70.4}                                                          & \multicolumn{3}{c}{91.9}                                      \\ \hline
\multicolumn{1}{l|}{GaitFormer~\cite{CVPR_23_MMGaitFormoer}} & \multicolumn{1}{c|}{GaitFormer} & \multicolumn{3}{c|}{53.4}                                                          & \multicolumn{3}{c}{85.3}                                      \\ \hline
\multicolumn{8}{c}{\textbf{CASIA-E~\cite{song2022casiae}}}                                                                                                                                                                                             \\ \hline
\multicolumn{2}{c|}{}                                             & \multicolumn{2}{c|}{NM}                               & \multicolumn{2}{c|}{BG}                                & \multicolumn{2}{c}{CL}            \\ \hline
\multicolumn{1}{l|}{GaitSet~\cite{gaitsetv1}}    & \multicolumn{1}{c|}{AAAI19}     & \multicolumn{2}{c|}{82.54}                            & \multicolumn{2}{c|}{75.26}                             & \multicolumn{2}{c}{62.53}         \\ \hline
\multicolumn{1}{l|}{GaitPart~\cite{gaitpart}}   & \multicolumn{1}{c|}{CVPR20}     & \multicolumn{2}{c|}{82.92}                            & \multicolumn{2}{c|}{74.36}                             & \multicolumn{2}{c}{60.48}         \\ \hline
\multicolumn{1}{l|}{GaitBase~\cite{fan2023opengait}}   & \multicolumn{1}{c|}{CVPR23}     & \multicolumn{2}{c|}{91.59}                            & \multicolumn{2}{c|}{86.65}                             & \multicolumn{2}{c}{74.73}         \\ \hline
\multicolumn{8}{c}{\textbf{CCPG~\cite{li2023CCPG}}}                                                                                                                                                                                               \\ \hline
\multicolumn{2}{c|}{\multirow{2}{*}{}}                            & \multicolumn{2}{c|}{RGB w\/o face}                     & \multicolumn{2}{c|}{RGB w\/o face \& foot}               & \multicolumn{2}{c}{Silhouettes}   \\ \cline{3-8} 
\multicolumn{2}{c|}{}                                             & \multicolumn{1}{c|}{R@1}  & \multicolumn{1}{c|}{mAP}  & \multicolumn{1}{c|}{rank1} & \multicolumn{1}{c|}{mAP}  & \multicolumn{1}{c|}{rank1} & mAP  \\ \hline
\multicolumn{1}{l|}{AP3D~\cite{li2023CCPG}}       & \multicolumn{1}{c|}{ECCV20}           & \multicolumn{1}{c|}{86.7} & \multicolumn{1}{c|}{60.1} & \multicolumn{1}{c|}{55.1}  & \multicolumn{1}{c|}{27.3} & \multicolumn{1}{c|}{-}     & -    \\ \hline
\multicolumn{1}{l|}{BiCnet-TKS~\cite{li2023CCPG}} & \multicolumn{1}{c|}{CVPR21}           & \multicolumn{1}{c|}{84.2} & \multicolumn{1}{c|}{57.9} & \multicolumn{1}{c|}{64.5}  & \multicolumn{1}{c|}{36.9} & \multicolumn{1}{c|}{-}     & -    \\ \hline
\multicolumn{1}{l|}{PSTA~\cite{li2023CCPG}}       & \multicolumn{1}{c|}{ICCV21}           & \multicolumn{1}{c|}{88.2} & \multicolumn{1}{c|}{65.3} & \multicolumn{1}{c|}{62.6}  & \multicolumn{1}{c|}{37.6} & \multicolumn{1}{c|}{-}     & -    \\ \hline
\multicolumn{1}{l|}{PiT~\cite{li2023CCPG}}        & \multicolumn{1}{c|}{TII22}           & \multicolumn{1}{c|}{85.1} & \multicolumn{1}{c|}{60.1} & \multicolumn{1}{c|}{57.1}  & \multicolumn{1}{c|}{30.8} & \multicolumn{1}{c|}{-}     & -    \\ \hline
\multicolumn{1}{l|}{GaitSet~\cite{gaitsetv1}}    & \multicolumn{1}{c|}{AAAI19}     & \multicolumn{1}{c|}{-}    & \multicolumn{1}{c|}{-}    & \multicolumn{1}{c|}{-}     & \multicolumn{1}{c|}{-}    & \multicolumn{1}{c|}{77.7}  & 46.4 \\ \hline
\multicolumn{1}{l|}{GaitPart~\cite{gaitpart}}   & \multicolumn{1}{c|}{CVPR20}     & \multicolumn{1}{c|}{-}    & \multicolumn{1}{c|}{-}    & \multicolumn{1}{c|}{-}     & \multicolumn{1}{c|}{-}    & \multicolumn{1}{c|}{77.8}  & 45.5 \\ \hline
\multicolumn{1}{l|}{GaitGL~\cite{Lin_2021_ICCV}}     & \multicolumn{1}{c|}{ICCV21}     & \multicolumn{1}{c|}{-}    & \multicolumn{1}{c|}{-}    & \multicolumn{1}{c|}{-}     & \multicolumn{1}{c|}{-}    & \multicolumn{1}{c|}{69.1}  & 27   \\ \hline
\end{tabular}
}
\end{table}
\subsubsection{Evaluation on clothes-changing gait recognition}
\pending{FVG~\cite{zhang2020gaitnet}, CASIA-E~\cite{song2022casiae}, and CCPG~\cite{li2023CCPG} are three up-to-date benchmarks for clothes-changing gait recognition. Among them, FVG~\cite{zhang2020gaitnet} is the smallest dataset consisting of 2,856 front-view walking sequences from 226 identities, and FVG also has a clothes-changing subset. CASIA-E~\cite{song2022casiae} contains 1,014 people and 778,752 videos. Each participant involves varied appearances caused by changes in carrying and dressing. However, it only provides silhouettes and infrared thermal images. CCPG~\cite{li2023CCPG} provides 200 identities and over 16K sequences which are captured indoors and outdoors, while each identity has seven different cloth-changing statuses. Besides, both CCPG and FVG provide raw RGB images. As shown in Table~\ref{tab:clothesdataset}, we observe that state-of-the-art methods can achieve over 70\% recognition under clothes-changing settings in all three datasets, and deep models fed with RGB images as input can gain significant performance improvement. However, CCPG addressed that the performance improvement may come from utilizing gait-irrelevant information such as face and shoes. We suggest it is potential to study gait recognition from RGB images in an end-to-end manner, as RGB-based methods are efficient and contain more information.}

\subsubsection{Evaluation on 3D gait recognition}
\pending{Recently, the integration of LiDAR sensors in gait recognition~\cite{shen2023lidargait,ahn20222vgait} has gained attention to improve human perception in challenging lighting conditions and provide precise 3D information. Notably, datasets like SUSTech1K~\cite{shen2023lidargait} and LiCamGait~\cite{han2022licamgait} have been introduced, offering a large-scale collection of sequences with various variations. Unlike silhouette-based methods that struggle with poor illumination, LiDAR-based gait recognition~\cite{shen2023lidargait,ahn20222vgait} demonstrates promising performance in such conditions, as well as in scenarios with occlusion and other realistic challenges. As shown in Table~\ref{tab:lidardataset}, the 3D geometry information provided by LiDAR sensors allows for the learning of more informative and discriminative gait representations, enhancing the practical applications of gait recognition.}

\begin{table}[htbp]
\label{tab:lidardataset}
\caption{\pending{Gait recognition performance on two LiDAR-based datasets, LiCamGait and SUSTech1K.}}
\scalebox{0.75}{
\begin{tabular}{lccccccccc}
\hline
\multicolumn{10}{c}{\textbf{LiCamGait~\cite{han2022licamgait}}}                                                                                                                                                                                                                                                   \\ \hline
\multicolumn{2}{c|}{\multirow{2}{*}{}}                               & \multicolumn{4}{c|}{2-8m}                                                               & \multicolumn{3}{c}{8-15m}                                                        &      \\ \cline{3-10} 
\multicolumn{2}{c|}{}                                                & \multicolumn{1}{c|}{NM}    & \multicolumn{1}{c|}{BG}   & \multicolumn{1}{c|}{CL}   & \multicolumn{1}{c|}{Mean} & \multicolumn{1}{c|}{NM}   & \multicolumn{1}{c|}{BG}   & \multicolumn{1}{c|}{CL}   & Mean \\ \hline
\multicolumn{1}{l|}{PointNet~\cite{qi2017pointnet}}         & \multicolumn{1}{c|}{CVPR17}  & \multicolumn{1}{c|}{32.3}  & \multicolumn{1}{c|}{29}   & \multicolumn{1}{c|}{44.8} & \multicolumn{1}{c|}{35.4} & \multicolumn{1}{c|}{29}   & \multicolumn{1}{c|}{19.4} & \multicolumn{1}{c|}{34.4} & 27.6 \\ \hline
\multicolumn{1}{l|}{PointMLP~\cite{ma2022pointmlp}}         & \multicolumn{1}{c|}{ICLR22}  & \multicolumn{1}{c|}{51.6}  & \multicolumn{1}{c|}{38.7} & \multicolumn{1}{c|}{55.1} & \multicolumn{1}{c|}{48.5} & \multicolumn{1}{c|}{35.5} & \multicolumn{1}{c|}{29}   & \multicolumn{1}{c|}{44.8} & 36.4 \\ \hline
\multicolumn{1}{l|}{Han~\etal~\cite{han2022licamgait}}         & \multicolumn{1}{c|}{Arxiv22} & \multicolumn{1}{c|}{77.42} & \multicolumn{1}{c|}{61.3} & \multicolumn{1}{c|}{68.8} & \multicolumn{1}{c|}{69.2} & \multicolumn{1}{c|}{54.8} & \multicolumn{1}{c|}{71}   & \multicolumn{1}{c|}{58.6} & 61.5 \\ \hline
\multicolumn{10}{c}{\textbf{SUSTech1K~\cite{shen2023lidargait}}}                                                                                                                                                                                                                                                   \\ \hline
\multicolumn{2}{c|}{}                                                & \multicolumn{1}{c|}{NM}    & \multicolumn{1}{c|}{BG}   & \multicolumn{1}{c|}{CL}   & \multicolumn{1}{c|}{CR}   & \multicolumn{1}{c|}{UB}   & \multicolumn{1}{c|}{UF}   & \multicolumn{1}{c|}{OC}   & NT   \\ \hline
\multicolumn{1}{l|}{PointNet~\cite{qi2017pointnet}}         & \multicolumn{1}{c|}{CVPR17}  & \multicolumn{1}{c|}{43.6}  & \multicolumn{1}{c|}{37.3} & \multicolumn{1}{c|}{25.7} & \multicolumn{1}{c|}{28.8} & \multicolumn{1}{c|}{19.9} & \multicolumn{1}{c|}{30.1} & \multicolumn{1}{c|}{44.3} & 27.4 \\ \hline
\multicolumn{1}{l|}{PointNet++~\cite{qi2017pointnet++}}       & \multicolumn{1}{c|}{NIPS17}  & \multicolumn{1}{c|}{55.9}  & \multicolumn{1}{c|}{52.2} & \multicolumn{1}{c|}{41.6} & \multicolumn{1}{c|}{49.6} & \multicolumn{1}{c|}{47.8} & \multicolumn{1}{c|}{45.9} & \multicolumn{1}{c|}{54.2} & 52.5 \\ \hline
\multicolumn{1}{l|}{SimpleView~\cite{goyal2021simpleview}}       & \multicolumn{1}{c|}{ICML21}  & \multicolumn{1}{c|}{72.3}  & \multicolumn{1}{c|}{68.8} & \multicolumn{1}{c|}{57.2} & \multicolumn{1}{c|}{63.3} & \multicolumn{1}{c|}{49.2} & \multicolumn{1}{c|}{62.5} & \multicolumn{1}{c|}{79.7} & 66.5 \\ \hline
\multicolumn{1}{l|}{LidarGait~\cite{shen2023lidargait}}        & \multicolumn{1}{c|}{CVPR23}  & \multicolumn{1}{c|}{91.8}  & \multicolumn{1}{c|}{88.6} & \multicolumn{1}{c|}{74.6} & \multicolumn{1}{c|}{89}   & \multicolumn{1}{c|}{67.5} & \multicolumn{1}{c|}{80.9} & \multicolumn{1}{c|}{94.5} & 90.4 \\ \hline
\end{tabular}
}
\end{table}

\begin{table*}[htbp]
\centering
\caption{All public gait datasets we can find in the literature. The related information of each dataset are also listed.}
\label{tab:dataset}
\scalebox{0.85}{
\begin{tabular}{llrrrllll} 
\hline
Institution                      & Dataset           & Subjects & Sequences & Views & Variations                                                                                                                                       & Environment & Available & Year  \\ 
\hline
SUSTech,China                        & SUSTech1K~\cite{shen2023lidargait}     & 1,050    & 25,239        & 12    & \begin{tabular}[c]{@{}l@{}}views, occlusion,\\ clothing, illumination,\\ 3D point cloud\end{tabular}                                                                                                                                & outdoor     & yes       & 2023  \\  
\hline
BJTU,China                        & CCPG~\cite{li2023CCPG}     & 200    & 16,566        & 10    & clothing                                                                                                                                & outdoor     & yes       & 2023  \\  
\hline
ZJU,China                        & VersatileGait~\cite{dataset2021versatilegait}     & 10,000    & 1,320,000        & 44  & age, gender, walking style                                                                                                                               & Unity3D     & yes       & 2021  \\ 
\hline
HDU,China                        & Gait3D~\cite{zheng2022gait3d}     & 4,000    & 253,309        & 39    & views                                                                                                                                 & in/outdoor     & yes       & 2021  \\ 
\hline
THU,China                        & GREW~\cite{dataset2021grew}              & 26,345    & 128,671    & 882   & \begin{tabular}[c]{@{}l@{}}view, distractor,~\\carrying, dressing,\\occlusion, surface,\\illumination, speed, \\shoes, trajectories\end{tabular} & wild        & yes       & 2021  \\ 
\hline
\multirow{2}{*}{SZU,China}       & ReSGait~\cite{dataset2021resgait}            & 172      & 870       & 1     & \begin{tabular}[c]{@{}l@{}}cl, carrying, \\trajectories\end{tabular}                                                                             & indoor      & yes       & 2021  \\ 
\cline{6-6}
                                 & RGB-D Gait~\cite{dataset2013szurgbd}        & 99       & 792       & 2     & views                                                                                                                                            & indoor      & yes       & 2013  \\ 
\hline
\multirow{13}{*}{OU-ISIR, Japan} & OUMVLP Pose~\cite{2020anPoseDataset}       & 10,307    & 268,086    & 14    & views                                                                                                                                            & indoor      & yes       & 2020  \\
                                 & OU-LP Bag~\cite{dataset2017oulpbag}         & 62,528    & 177,973    & 1     & carrying                                                                                                                                         & indoor      & yes       & 2018  \\
                                 & OUMVLP~\cite{dataset2017oumvlp}            & 10,307    & 267,386    & 14    & views                                                                                                                                            & indoor      & yes       & 2018  \\
                                 & OU-LP Age~\cite{dataset2017oulpage}          & 63,846    & 63,846     & 30    & age                                                                                                                                              & indoor      & yes       & 2017  \\
                                 & Bag $\beta$~\cite{dataset2018oulpbagb} & 2,070     & 4,140      & 1     & carrying                                                                                                                                          & indoor      & yes       & 2017  \\
                                 & ST-1~\cite{dataset2014ou_st1}              & 179      & -         & 1     & speed                                                                                                                                            & indoor      & yes       & 2014  \\
                                 & ST-2~\cite{dataset2014ou_st1}              & 178      & -         & 1     & speed                                                                                                                                            & indoor      & yes       & 2014  \\
                                 & OU-LP c1v1~\cite{dataset2012oulpc1v1}        & 4,007     & 7,844      & 1     & -                                                                                                                                                & indoor      & yes       & 2012  \\
                                 & OU-LP c1v2~\cite{dataset2012oulpc1v2}        & 4,016     & 7,860      & 1     & -                                                                                                                                                & indoor      & yes       & 2012  \\
                                 & Speed~\cite{dataset2012ou_treadmill_scvf}             & 34       & 612       & 1     & speed                                                                                                                                            & indoor      & yes       & 2012  \\
                                 & clothing~\cite{dataset2012ou_treadmill_scvf}          & 68       & 2,746      & 1     & clothing                                                                                                                                         & indoor      & yes       & 2012  \\
                                 & view~\cite{dataset2012ou_treadmill_scvf}              & 200      & 5,000      & 1     & views                                                                                                                                            & indoor      & -         & 2012  \\
                                 & fluctuation~\cite{dataset2012ou_treadmill_scvf}       & 185      & 370       & 1     & fluctuation                                                                                                                                      & indoor      & yes       & 2012  \\ 
\hline
UMA, Spain                       & MuPeG~\cite{dataset2020umaMuPeG}             & -        & -         & -     & occlusion                                                                                                                                        & indoor      & yes       & 2020  \\ 
\hline
Kyudai, Japan                       & PCG~\cite{yamada2020PCG_pcdataset}             & 30        & 60         & 1     & 3D point cloud                                                                                                                                        & -      & yes       & 2020  \\ 
\hline
MSU, US                          & FVG~\cite{zhang2019gait}               & 226      & 2,856      & 3     & \begin{tabular}[c]{@{}l@{}}views, speed,\\carrying,~cl\end{tabular}                                                                  & outdoor     & yes       & 2019  \\ 
\hline
IPVC, Portugal                   & GRIDDS~\cite{dataset_ipvc_gridds,zhang2020gaitnet}            & 35       & 350       & 1     & trajectories                                                                                                                                     & indoor      & yes       & 2019  \\ 
\hline
ISR-Lisboa, Portugal             & ks20              & 20       & 300       & 5     & view                                                                                                                                             & indoor      & -         & 2017  \\ 
\hline
GPJATK             & GPJATK              & 32       & 166       & 4     & view, 3D data                                                                                                                                             & indoor      & -         & 2017  \\ 
\hline
SDU, China                       & SDUGait~\cite{dataset_sdu}           & 52       & 1,040      & 2     & trajectories views                                                                                                                               & indoor      & yes       & 2016  \\ 
\hline
MTA–SZTAKI, Hungary                       & SZTAKI-LGA~\cite{benedek2016lidar_dataset}           & 28       & 11      & 1     & 3D point cloud                                                                                                                               & outdoor      & yes       & 2016  \\ 
\hline
WUST, Polan                      & BHV MoCap~\cite{dataset_wust_mocap}         & 10       & 246       & 1     & trajectories                                                                                                                                     & -           & yes       & 2015  \\ 
\hline
IITs, Indian                     & Depth Gait~\cite{dataset_iits_depthGait}       & 29       & 464       & 2     & \begin{tabular}[c]{@{}l@{}}view, occlusion, \\speed\end{tabular}                                                                                 & -           & yes       & 2015  \\ 
\hline
PPGC-UFPel, Brasil               & Kinect~\cite{dataset_ppgc_kinect}            & 164      & 820       & -     & curve                                                                                                                                            & indoor      & yes       & 2015  \\ 
\hline
\multirow{3}{*}{KY, Japan}       & KY4D-B~\cite{dataset_kyb}            & 42       & 84        & 16    & curve                                                                                                                                            & indoor      & yes       & 2014  \\
                                 & Shadow~\cite{dataset_kyshadow}            & 54       & 324       & 1     & views, cl, bg                                                                                                                                    & indoor      & yes       & 2014  \\
                                 & KY4D-A~\cite{dataset_kya}            & 42       & 168       & 16    & views                                                                                                                                            & indoor      & yes       & 2010  \\ 
\hline
A.V.A UCO, Spain                 & AVAMVG~\cite{dataset_AVAMVG}            & 20       & 1,200      & 6     & views, trajectories                                                                                                                              & indoor      & yes       & 2013  \\ 
\hline
WVU, US                          & WOSG~\cite{dataset_wosg}              & 155      & -         & 8     & views, Illumination                                                                                                                              & outdoor     & -         & 2013  \\ 
\hline
ITB, Indonesian                  & dataset~\cite{dataset_indonesian}           & 212      & -         & 1     & -                                                                                                                                                & indoor      & -         & 2012  \\ 
\hline
\multirow{2}{*}{TUM, Germany}             & TUM-IITKGP~\cite{dataset_tum_iitkgp}        & 305      & 3,370      & 1     & \begin{tabular}[c]{@{}l@{}}time, carrying, \\shoes, occlusion\end{tabular}                                                                                  & Indoor      & yes        & 2012  \\ 
\cline{6-6}
                                 & TUM-GAID~\cite{dataset_tum_gaid}          & 35       & 1,645      & 1     & \begin{tabular}[c]{@{}l@{}}times, appearance, \\bg, depth\end{tabular}                                                                     & indoor      & no       & 2010  \\ 
\hline
UAB, Spain                       & DGait~\cite{dataset_DGAIT}             & 55       & 605       & 1     & trajectories                                                                                                                                     & indoor      & yes       & 2012  \\ 
\hline
IIT, Italy                       & RGBD-ID~\cite{dataset_iit_rgbdid}           & 79       & 316       & 1     & \begin{tabular}[c]{@{}l@{}}trajectories, time,\\~cl, speed\end{tabular}                                                                          & indoor      & yes       & 2012  \\ 
\hline
QUT, Australia                   & SAIVT-DGD~\cite{dataset_qut_saivt-dgd}         & 35       & 700       & 1     & \begin{tabular}[c]{@{}l@{}}speed, carrying, \\shoes\end{tabular}                                                                                 &             & yes       & 2011  \\ 
\hline
\multirow{5}{*}{Soton, England}  & Multimodal~\cite{dataset_soton_multimodoality_1,dataset_survey_makihara}        & 300      & 5,000      & 12    & views                                                                                                                                            & indoor      & yes       & 2011  \\
                                 & Temporal~\cite{dataset_soton_temporal}          & 25       & 2,280      & 12    & views                                                                                                                                            & indoor      & yes       & 2011  \\ 
\cline{6-6}
                                 & Small~\cite{nixonbook}             & 12       & -         & 4     & \begin{tabular}[c]{@{}l@{}}bg,cl,carrying, \\ speed, footwear, \\views\end{tabular}                                                              & indoor      & yes       & 2002  \\ 
\cline{6-6}
                                 & Large~\cite{dataset_soton_large}             & 116      & 2,128      & 2     & \begin{tabular}[c]{@{}l@{}}Terrain, direction, \\views\end{tabular}                                                                              & in/outdoor  & yes       & 2002  \\
                                 & Early~\cite{nixonbook}             & 10       & 40        & 1     & -                                                                                                                                                & indoors     & -         & 1997  \\ 
\hline
TIT, Japan                       & TokyoTech DB~\cite{dataset_tit_tokyotech}      & 30       & 1,902      & -     & speed                                                                                                                                            & indoor      & -         & 2010  \\ 
\hline
\multirow{4}{*}{CASIA, China}    & CASIA-D~\cite{dataset_casiad}           & 88       & 880       & 1     & multi-modality                                                                                                                                   & indoor      & yes       & 2009  \\
                                 & CASIA-B~\cite{casiab}           & 124      & 13,640     & 11    & views cl bg                                                                                                                                      & indoor      & yes       & 2005  \\
                                 & CASIA-C~\cite{dataset_casiac}           & 153      & 1,530      & 1     & speed bg                                                                                                                                         & outdoor     & yes       & 2005  \\
                                 & CASIA-A~\cite{dataset_wang2003silhouette}           & 20       & 240       & 1     & walking direction                                                                                                                                & outdoor     & yes       & 2001  \\ 
\hline
BUAA, China                      & IRIP              & 60       & 4,800      & 8     & gender view                                                                                                                                      & indoor      & -         & 2008  \\ 
\hline
GT, US                           & GT Speed~\cite{dataset_gatech}          & 20       & 268       & 3     & views time                                                                                                                                       & in/outdoor  & yes       & 2003  \\ 
\hline
USF, US                          & USF~\cite{usfgait,usflabel}               & 122      & 1,870      & 2     & \begin{tabular}[c]{@{}l@{}}shoes, views, \\carrying, terrain,\\time, trajectories\end{tabular}                                                   & outdoor     & yes       & 2002  \\ 
\hline
\multirow{3}{*}{UMD, US}         & Dataset-1~\cite{dataset_umd_online}         & 25       & 100       & 4     & views, long distance                                                                                                                             & outdoor     & yes       & 2001  \\
                                 & Dataset-2~\cite{dataset_umd_online}         & 55       & 222       & 2     & views, times                                                                                                                                     & outdoor     & yes       & 2001  \\
                                 & Dataset-3~\cite{dataset_umd_online}         & 12       & -         & 1     & views                                                                                                                                            & outdoor     & yes       & 2001  \\ 
\hline
CMU, US                          & CMU-mobo~\cite{dataset_cmu_mobo}          & 25       & 600       & 6     & \begin{tabular}[c]{@{}l@{}}speed, carrying,\\inclination\end{tabular}                                                                            & indoor      & yes       & 2001  \\ 
\hline
\multirow{2}{*}{MIT, US}         & MITAI Gait~\cite{dataset_mit_ai}        & 24       & 194       & 1     & times months                                                                                                                                     & indoors     & yes       & 2001  \\
                                 & Early~\cite{dataset_mit_early}             & 5        & 26        & 1     & -                                                                                                                                                & indoor      & -         & 1994  \\ 
\hline
UCSD, US                         & UCSD~\cite{dataset_ucsd}              & 6        & 42        & 1     & -                                                                                                                                                & outdoor     & yes       & 1998  \\ 
\hline
NTTBRL, Japan                    & NIT Gait~\cite{dataset_nit_gait}          & 7        & 70        & 1     & same cl, shoes                                                                                                                                   & -           & -         & 1995  \\
\hline
\end{tabular}}
\end{table*}

\section{Security and Privacy of Gait Recognition}
The rapid development of gait recognition raises concerns that both the research community and society at large should address the potential effects. While the overview on biometrics by Jain et al.~\cite{jain2021tbiom} provides a comprehensive description of security and privacy, it covers the broader field of biometrics and does not specifically focus on gait recognition. To bridge this gap, we will summarize the security and privacy challenges in biometrics in general and emphasize the specific concerns related to gait recognition.

\subsection{Security}
\label{sec:security}
Like other biometrics systems, gait recognition should also be secure from various attacks. There are three kinds of attacks, according to the summary in~\cite{jain2021tbiom}, presentation attacks, adversarial attacks, and template attacks.

\begin{itemize}

\item \textbf{Presentation Attacks} refer to a type of attack where artificial objects are presented to the sensors of biometric systems. This form of attack is prevalent in face recognition systems, as highlighted in the FacePAD study~\cite{facepad}, where face images displayed on devices or 3D silicon masks are used to deceive the system. In contrast, there have been limited studies on gait presentation attacks. The pioneering investigation on vision-based gait presentation attacks was conducted by Hadid~\etal~\cite{2012spoofnixon}. However, presentation attacks in gait recognition remain relatively unexplored, necessitating further research in this area.

\item \textbf{Adversarial Attacks} are by digital synthetic data. Unlike presentation attacks, which rely on physical objects, adversarial attacks are based on generative models that can synthesize realistic data~\cite{yu2017gaitgan}. The advancement of generative models, such as stable diffusion models, has enabled the creation of high-quality and lifelike faces and human figures through extensive training on large datasets. While it is feasible to generate single frames that preserve identity with high fidelity, gait recognition systems require sequential data, necessitating both fidelity and temporal consistency. Further research is needed to explore these two critical aspects of fidelity and consistency in the context of gait recognition to enhance its robustness against adversarial attacks.

\item \textbf{Template attacks} is to reconstruct images or videos from templates that are extracted by a biometrics system. Studies on face~\cite{facetemplateattack} have shown its feasibility. Unlike presentation and adversarial attacks, template attacks are manipulated at the feature level so that gait template attacks can borrow insights from other biometric features such as the face, fingerprint, or iris. 

\end{itemize}

\subsection{Privacy}
\label{sec:privacy}
Gait recognition systems, as their significant performance achieved in both indoor and outdoor scenarios, may pose greater privacy concerns compared to face recognition systems, which have already raised significant privacy issues globally. Gait can be captured from a greater distance than facial features. While it is natural to wear hats, sunglasses, or masks to protect one's face, concealing gait through completely different clothing is not as practical. In addition to identity, gait can also reveal information about gender and health conditions, as demonstrated in studies~\cite{yu2008gender}. These factors contribute to the increased privacy implications associated with gait recognition systems.

To protect data and privacy, various regulations and laws have been released worldwide. The European Parliament introduced the \textit{General Data Protection Regulation (GDPR)} in 2016, and \textit{Data Security Law of the P. R. of China}~\cite{law-dsl} went into effect in 2021. Similarly, Several states in the US also passed similar laws, such as \textit{California Consumer Privacy Act}~\cite{ccpa} and \textit{Biometric Information Privacy Act} of Illinois. In China, a national standard called \textit{Information Security Technology—Personal Information Security Specification (GB/T 35273-2020)}~\cite{PISS} has been passed, providing detailed instructions for personal information security. Most laws and regulations concerning the privacy protection of biometric data share common principles. They typically outline the rights of individuals whose biometric data is collected, including the right 1) to know how the data will be used and stored, 2) to delete the data, and 3) to opt-out of the data usage. Meanwhile, it is worth noting that many laws primarily impose restrictions on the use of biometric data by private businesses, while allowing certain exceptions for government agencies for purposes such as public security. This aspect of the laws, which pertains to the usage of biometric data by governments, often lacks transparency and may raise concerns regarding the extent and potential misuse of such data for surveillance or other undisclosed purposes.

\begin{figure}[htbp]
\centering
\includegraphics[width=0.48\textwidth]{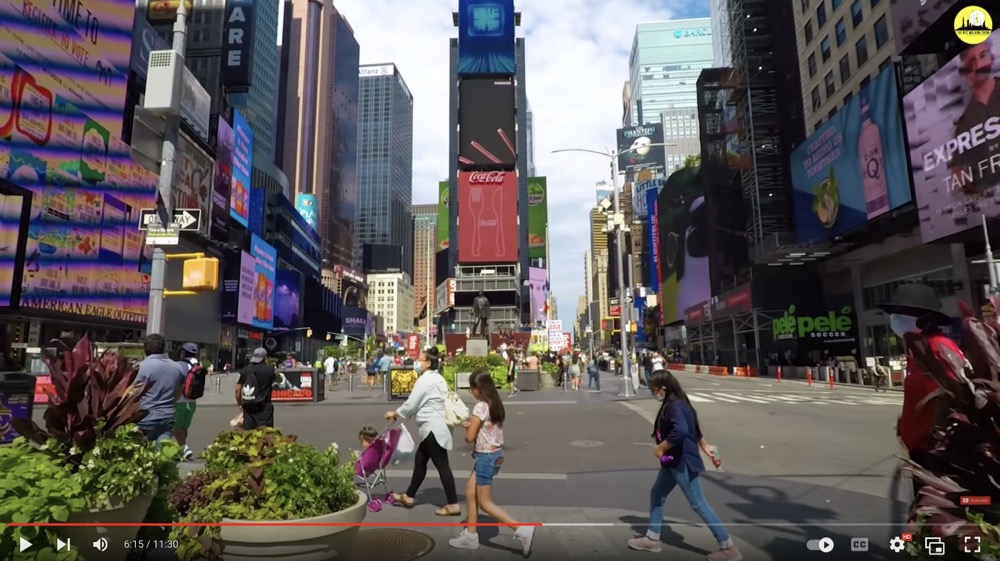}
\caption{A shared video on the social media platform with many pedestrians~\cite{streetwalking-youtube}.}
\label{fig:streetwalking}
\end{figure}

The privacy concerns on gait recognition may bring a crisis to video-sharing social platforms such as YouTube and TikTok. There are many kinds of videos, as in Fig.~\ref{fig:streetwalking} online. If the walking pedestrians in the videos can be identified by their gait features, should we get a permit from them before posting? It is impossible to get permits from all pedestrians. Google Street has blurred all human faces on the street. Should the pedestrians in the videos be blurred also, or should those videos be deleted directly from the Internet? We leave these questions open for the community and society to discuss.

\section{Challenges and Directions} 
Even though significant progress has been achieved in the past years, there are still many challenges in gait recognition. We think gait recognition can be improved from datasets, practicality, and trustworthiness. They are described in detail in the following part of this section.

\subsection{Datasets}

\noindent$\bullet$~\textit{Learning from real scenarios.}
Gait recognition using deep learning techniques has achieved promising results in indoor datasets captured in constrained scenarios. Although indoor laboratory datasets typically contain limited identities, these datasets help address many challenges of gait recognition, such as changes in clothes or viewpoints. Recently many outdoor datasets have been proposed to promote the study of gait recognition in unconstrained scenarios and extend the recognition research to realistic factors like poor illumination and low resolutions~\cite{dataset2021grew,shen2023lidargait}. In-the-wild datasets~\cite{dataset2021grew,zheng2022gait3d,li2023CCPG,shen2023lidargait} help develop a promising gait recognition system. However, collecting a large-scale gait dataset can be extremely difficult as gait is ambiguous to annotate over different viewpoints. The ambiguity of gait really prevents us from establishing a large-scale gait dataset with diverse viewpoints, clothing, and status for each pedestrian, and it is worth investigating potential solutions to extend gait recognition in real scenarios.

\noindent$\bullet$~\textit{Learning from synthetic data.}
Since it is challenging to collect gait data in real scenarios, one possible solution is to use synthetic methods to generate gait data. Synthetic data is prevalent in many studies, including face spoofing, self-driving~\cite{syntheticselfdriving}, and 3D object classification. In gait recognition, VersatileGait~\cite{dataset2021versatilegait} is the only public, synthetic dataset. It is convenient and budget-friendly to obtain both fruitful attributes and large-scale identities. However, defining an identity may be a problem. There are many parameters to define a 3D body and its motion. If we consider two virtual persons to be two different identities, what will be the threshold of their similarity?

\noindent$\bullet$~\textit{Learning from unlabelled data.}
Collecting a large gait dataset in real scenarios is difficult and potentially causes privacy concerns. Another possible solution is to use unlabelled data. We can collect many gait data from videos online and other sources~\cite{fan2022gaitlu}. Gait data can be automatically detected and segmented from videos. Although it is difficult to label the identities in those videos, deep models still have the potential to learn informative representations from unlabelled data via self-supervised learning methods. We value semi- and self-supervised learning methods to promote the study of gait recognition in the future greatly.

\subsection{Gait Recognition Toward Reality}

\pending{\noindent$\bullet$~\textit{Gait-Changing Gait Recognition.}
Several common factors can significantly impact gait and result in performance degradation in gait recognition systems. Camera viewpoints, variations in pedestrian dressing, and occlusion are examples of factors that can lead to significant differences in human appearance, making it challenging to match samples from the same subject successfully. To tackle this challenge, researchers have explored potential solutions such as eliminating gait-irrelevant information through the disentanglement method~\cite{zhang2020gaitnet} or modeling gait motion information to be robust to appearance changes~\cite{ma2023dynamic_CVPR,dou2022metagait}. However, it is important to note that gait motion itself can also vary due to factors such as aging and disease. This aspect is often overlooked but presents valuable opportunities for further exploration in the field of gait recognition.}

\pending{\noindent$\bullet$~\textit{End-to-end Gait Recognition.}
The current focus of gait recognition has predominantly been on the recognition process itself, ignoring the importance of upstream tasks that provide gait representations. These upstream tasks include human foreground segmentation and pose estimation methods, which are crucial for obtaining silhouettes and skeletons. However, recent research has highlighted that using off-the-shelf segmentation methods can negatively impact representation learning. As a result, there is a growing trend toward learning deep gait representations in an end-to-end manner. This approach involves applying deep models directly to raw input data, such as in the works of Liang~\etal~\cite{liang2022gaitedge}, Song~\etal~\cite{song2019gaitnet}, and Zhang~\etal~\cite{zhang2020gaitnet}. The objective is to fully leverage the whole-body information from the raw input, simplifying the overall pipeline of human identification.}

\pending{\noindent$\bullet$~\textit{Heterogeneous Gait Recognition.}
Most gait recognition methods primarily rely on camera-based modalities, which restricts their effectiveness in poor illumination conditions, particularly during nighttime. However, recent advancements in the field have introduced alternative modalities, including infrared cameras and LiDAR sensors, to overcome the limitations of camera-based approaches. These new modalities not only offer a promising solution for low-light gait recognition but also present new challenges, such as multimodal fusion and cross-modal retrieval, which require further investigation in the coming years.}

\pending{\noindent$\bullet$~\textit{Efficient Gait Recognition.}
The existing gait recognition methods have shown satisfactory performance on large-scale datasets~\cite{dataset2021grew}. However, they still fall short of meeting the requirements of practical surveillance systems, which demand real-time feature extraction and retrieval. The importance of efficiency in real-world applications is often overlooked~\cite{yolo}. In other fields of study, advancements have been made in lightweight models that offer comparable performance with reduced computational costs. Additionally, the existing literature fails to address the challenges posed by real-world applications that may involve galleries with millions or even billions of samples. The process of comparing a probe feature with such a large gallery can be extremely time-consuming. Therefore, there is a need to explore further and develop efficient gait recognition techniques that can meet the demands of practical applications.}

\pending{\noindent$\bullet$~\textit{Transfer learning for Gait Recognition.} 
Transfer learning in the context of gait recognition holds significant value and potential for various applications. It offers the advantage of training a model on a large dataset in one domain and then applying the learned knowledge to a different domain for evaluation. Two examples illustrate the benefits of transfer learning. Firstly, by training a deep model on a source domain like GREW~\cite{dataset2021grew}, we can leverage the learned knowledge to achieve promising results on a targeted domain such as CASIA-B~\cite{casiab}. This transfer of knowledge helps in improving performance and generalization~\cite{selfgait2021}. Additionally, transfer learning enables the utilization of learned knowledge from human gait recognition in animal recognition scenarios. By leveraging the insights gained from human gait recognition, we can potentially apply similar techniques and models to recognize and analyze the gaits of animals. Overall, transfer learning plays a vital role in gait recognition by enabling knowledge transfer across domains, leading to improved performance and the exploration of new application scenarios.}

\subsection{Trustworthy Gait Recognition}

\noindent$\bullet$~\textit{Gait Privacy Protection.} 
In addition to identifying individuals, gait recognition also raises privacy concerns regarding sensitive information such as race, gender, age, dress, and other attributes, similar to face recognition~\cite{raceface2019iccv}. It is crucial to address these privacy concerns and investigate potential biases associated with these factors in gait recognition systems. Future research should focus on understanding and mitigating these privacy issues to ensure fairness, transparency, and responsible deployment of gait recognition technology.

\noindent$\bullet$~\textit{Gait Encryption for Security Protection}
As discussed in Section~\ref{sec:privacy}, the growing accuracy of gait recognition poses challenges for online video-sharing platforms. Blurring or masking all pedestrians in videos can significantly degrade the visual quality, discouraging users from sharing such content. To address this issue, one potential solution could be to encrypt gait videos using modifications that render them unrecognizable by gait recognition methods while preserving the visual appeal. However, research on this specific topic is currently lacking, despite the existence of encryption methods for face recognition~\cite{faceencrypt2021iccv}. Further exploration and investigation are needed to develop effective and practical encryption techniques for protecting privacy in gait recognition videos.

\section{Conclusions}
This survey provided a comprehensive overview of deep gait recognition, covering its fundamental concepts, challenges, advancements, and potential future directions. The survey highlighted the various approaches and techniques, including traditional and deep learning-based approaches. It also discussed the importance of datasets, challenges related to privacy and security, and the emergence of alternative modalities. Furthermore, the survey reviewed the available deep gait recognition methods from feature learning and architecture perspectives following our proposed taxonomy, which provides a clear and thorough understanding of the big picture of the field. Overall, we hope this survey can serve as a valuable resource for researchers and practitioners in the field of gait recognition, shedding light on its advancements and potential for further development.

\ifCLASSOPTIONcompsoc
    \section*{Acknowledgments}
\else
    \section*{Acknowledgment}
\fi
This work is supported in part by the National Natural Science Foundation of China under Grant 61976144, and in part by the National Key Research and Development Program of China (Grant No. 2020AAA0140002).

\ifCLASSOPTIONcaptionsoff
  \newpage
\fi

{\small
  \bibliographystyle{ieeetr}
  \bibliography{conferences_abrv,reference}
}

\end{document}